\documentclass{article}

\usepackage[preprint]{neurips_2026}

\usepackage[utf8]{inputenc}
\usepackage[T1]{fontenc}
\usepackage{hyperref}
\usepackage{url}
\usepackage{booktabs}
\usepackage{amsfonts}
\usepackage{amsmath}
\usepackage{amssymb}
\usepackage{nicefrac}
\usepackage{microtype}
\usepackage{xcolor}
\usepackage{graphicx}
\usepackage{float}
\usepackage{subcaption}
\usepackage{mdframed}
\graphicspath{{figures/}{./}}
\numberwithin{table}{section}
\numberwithin{figure}{section}
\usepackage{multirow}
\usepackage{array}
\usepackage{natbib}
\usepackage{lineno}

\title{LinAlg-Bench: A Forensic Benchmark Revealing Structural Failure Modes in LLM Mathematical Reasoning}
\author{%
  Shradha Agarwal \quad Deepak Rajbhar \quad Tariq J. \\
  Department of Computer Science \& Department of Nuclear Engineering and Radiation Science \\
  Missouri University of Science and Technology \\
  Rolla, Missouri \\
  \texttt{sabrc@mst.edu}
}

\newcommand{\profiletable}[1]{%
  \smallskip\resizebox{\textwidth}{!}{%
  \begin{tabular}{@{}llllllll@{}}
  \toprule
  \textbf{Dim.} & \textbf{Reading} & \textbf{Arithmetic} & \textbf{Sequential} & \textbf{Det} & \textbf{Eig} & \textbf{Overall} & \textbf{Dominant failure} \\
  \midrule #1
  \bottomrule\end{tabular}}\smallskip}
\begin{document}
\maketitle

\begin{abstract}
We introduce LinAlg-Bench, a diagnostic benchmark evaluating 10 frontier
large language models on structured linear algebra computation across a
strict dimensional gradient of $3\times3$, $4\times4$, and $5\times5$
matrices. Spanning 9 task types and 660 SymPy-certified problems, the
benchmark exhaustively evaluates 6,600 model outputs. Beyond binary
accuracy, LinAlg-Bench introduces a three-stage automated forensic
pipeline classifying 1,156 failures into ten primary error tags with
fine-grained subtypes, revealing that LLM mathematical failure is not
random but structurally constrained by algorithm type and matrix
dimension. Our central finding is a sharp behavioral threshold at
$4\times4$ scale: below it, models fail through execution errors ---
sign tracking failures, arithmetic drift, and parity errors; above it,
failure transitions to computational abandonment, with models fabricating
responses through tool roleplay, constraint-consistent confabulation, and
structured hallucination rather than attempting computation. This
fabrication-to-abandonment transition is near-universal across all model
tiers and architectures, suggesting a working memory limit rather than a
knowledge gap, supported by three scale-emergent error types absent at
$3\times3$ but present at $4\times4$ and $5\times5$. We further show
that solution strategy rigidity is a near-perfect predictor of $5\times5$
determinant accuracy, document constraint-aware confabulation as a novel
structured hallucination failure mode, and release all data, model
outputs, error labels, and judge pipeline publicly.
\end{abstract}

\section{Introduction}
\label{sec:intro}

Do frontier large language models reason mathematically, or do they
approximate the surface structure of mathematical discourse? The
prevailing account of LLM mathematical failure is essentially statistical
--- models fail because they have seen insufficient training examples, or
because the problem exceeds some general capability threshold. This paper
argues for a stronger, structurally grounded account: LLM failure under
recursive computational load is not random, it follows a predictable
pattern determined by algorithm family and matrix dimension, and
transitions sharply from execution failure to computational abandonment
at a measurable complexity threshold.

Linear algebra provides an ideal testbed because operations are
algorithmically precise, scale naturally across matrix dimensions without
changing the underlying task, and decompose into a cognitive hierarchy in
which each level adds exactly one computational requirement absent in the
level below: matrix reading, parallel arithmetic, sequential state
tracking, recursive sign management, and compositional operations built
on prior levels. A model that fails on a $5\times5$ determinant cannot
attribute the failure to unfamiliar mathematics --- the algorithm is
identical to the $3\times3$ case it solved correctly.

At $3\times3$, nine of ten models achieve perfect or near-perfect
determinant accuracy. At $5\times5$, only three models exceed 50\% on
determinants, and eigenvalue accuracy collapses to near-zero for all
models. This pattern --- preserved knowledge, collapsed execution --- is
the signature of a working memory constraint rather than a knowledge
deficit. The mode of failure also changes categorically: at $3\times3$,
models attempt computation and fail within it; at $5\times5$, models
abandon computation before attempting it, fabricating responses through
structured hallucination that satisfies surface mathematical constraints
while being fundamentally incorrect. This fabrication-to-abandonment
threshold, documented across 6,600 exhaustively evaluated outputs, is
the central empirical contribution of this paper.

Models separate cleanly into three tiers defined by the dimensional
boundary at which this threshold is crossed --- Tier~3 models collapse
at the Recursive level, Tier~2 partially withstands Recursive but
collapses at the Compositional level, Tier~1 sustains Recursive but
converges to near-zero eigenvalue accuracy by $5\times5$. A targeted
ablation study further shows that enforcing algorithmically efficient
strategies does not recover accuracy, establishing that the bottleneck
operates at the level of autoregressive execution depth rather than
method selection.

LinAlg-Bench contributes: 660 SymPy-verified problems across 9 tasks and
3 matrix dimensions with all 6,600 outputs exhaustively evaluated; a
three-stage forensic pipeline classifying 1,156 failures into ten error
tags validated against 593 human-annotated labels; the
fabrication-to-abandonment behavioral threshold as a novel, falsifiable
empirical finding; and constraint-aware confabulation as a structurally
distinct hallucination failure mode. All data, model outputs, error
labels, and judge pipeline are publicly released.

\section{Related Work}
\label{sec:related}

Mathematical reasoning evaluation has progressed from arithmetic word
problems \citep{cobbe2021gsm8k} through competition-level challenges
\citep{hendrycksmath2021}, unified multi-domain evaluations
\citep{mishra2022lila}, to symbolic mathematics \citep{lample2020deep}
and multi-level comprehension \citep{liu2024mathbench}. These benchmarks
establish difficulty gradients but measure only final-answer accuracy ---
a model failing on GSM8K and a model failing on a $5\times5$ determinant
are treated identically despite failing for categorically different
reasons. LinAlg-Bench addresses this gap by holding the algorithm fixed
and scaling only matrix dimension, isolating computational depth from
mathematical knowledge --- a design not available in benchmarks where
difficulty and novelty are confounded. This enables a controlled test of
whether failure mode, not merely failure rate, changes with task
complexity.

Transformer models solve multi-step compositional tasks via shallow
pattern-matching rather than genuine reasoning \citep{dziri2023faith},
cannot reliably self-correct arithmetic errors \citep{huang2023large},
and confabulate primarily through factual misgrounding
\citep{maynez2020faithfulness} or overconfident generation
\citep{kadavath2022language}. LinAlg-Bench reveals a structurally
distinct failure mode --- constraint-aware confabulation --- in which
models fabricate mathematically plausible responses under computational
overload: eigenvalue sums match the matrix trace in 45\% of
fabrications, and magnitudes respect the Frobenius norm bound in 85\% of
cases ($n{=}20$ Ungrounded\_Guess cases). A benchmark evaluating only
necessary conditions would systematically misclassify these fabrications
as correct. The tool roleplay collapse documented in
Section~\ref{sec:confabulation}, in which models simulate invoking
external tools they cannot access, represents a further structured
hallucination mechanism not previously documented.

The mechanistic interpretability literature has developed tools for
localising computation within transformer architectures
\citep{vig2019multiscale, meng2022locating, conmy2023automated}.
LinAlg-Bench contributes at the behavioural level: the forensic taxonomy
generates testable hypotheses for future mechanistic work ---
sign-tracking failures predict late-layer parity circuit involvement,
and Complete\_Collapse predicts suppression of those circuits rather than
their corruption. Full mechanistic validation is reserved for follow-up
work.

\section{Benchmark Design}
\label{sec:benchmark}

LinAlg-Bench organises 9 linear algebra tasks into five cognitive levels
based on the computational demands each operation places on a language
model's processing pipeline. This taxonomy is not a difficulty ranking
but a causal decomposition: each level adds exactly one computational
requirement absent in the level below (Appendix~\ref{app:tasks}). The
critical boundary falls between Sequential and Recursive: determinant
computation requires maintaining hierarchical parity state across nested
cofactor expansions simultaneously, while rank and nullity require only
independent row operations. Eigenvalue at $5\times5$ compounds this as a
deliberate stress test --- not a standard capability measurement ---
requiring a full $5\times5$ determinant followed by degree-5 polynomial
root-finding.

\begin{table}[t]
\caption{Five cognitive levels of the LinAlg-Bench taxonomy. Each level
adds exactly one computational requirement absent in the level below.
The critical boundary falls between Sequential and Recursive: row
operations are independent, while cofactor expansion requires maintaining
hierarchical parity state across nested submatrix expansions
simultaneously.}
\label{tab:taxonomy}
\centering
\begin{tabular}{p{2.2cm}p{3.8cm}p{3cm}p{3.8cm}}
\toprule
\textbf{Level} & \textbf{Cognitive Demand} & \textbf{Tasks} & \textbf{Key Operation} \\
\midrule
Reading       & Pure entry extraction                 & Trace, Transpose                            & Read and report matrix entries \\
Arithmetic    & Parallel independent computation      & Matrix-Vector, Multiplication, Matrix-Power & Independent dot products \\
Sequential    & State tracking through row operations & Rank, Nullity                               & Gaussian elimination with state updates \\
Recursive     & Deep recursive sign management        & Determinant                                 & Cofactor expansion with $\pm1$ parity tracking \\
Compositional & Built on Determinant as sub-routine   & Eigenvalues                                 & Characteristic polynomial via determinant \\
\bottomrule
\end{tabular}
\end{table}

The benchmark comprises 660 integer-entry problems (220 per dimension;
det=50, eig=30, others=20 per model), ground-truth verified against
SymPy symbolic computation \citep{meurer2017sympy}. Determinant and
eigenvalue receive larger sample allocations (50 and 30 questions
respectively) because they are the primary Recursive and Compositional
tasks under investigation; the remaining seven subcategories use
$n{=}20$, sufficient to establish ceiling and floor effects at Reading,
Arithmetic, and Sequential levels. Ten frontier models spanning
reasoning-optimised, mixture-of-experts, and standard instruction-tuned
architectures were evaluated zero-shot at temperature 0 via API; full
model details are in Appendix~\ref{app:models}. Code execution and
external tools were disabled --- this is a deliberate synthetic stress
test of unaided recursive execution under increasing depth, not a
practical evaluation of tool-use capability. Format sensitivity was
evaluated at $4\times4$ and $5\times5$ across three input representations
(LaTeX, Tabular, List) for nine models; format analysis targets
non-reasoning architectures where parsing load interacts with
computational depth.

\section{Results}
\label{sec:results}

Reading and Arithmetic level tasks --- trace, transpose, matrix-vector,
multiplication, matrix-power --- remain essentially flat across all three
dimensions, confirming that basic matrix comprehension and parallel
arithmetic place no meaningful load at any scale. Sequential tasks (rank,
nullity) show a gradual, monotonic decline consistent with increasing
\textbf{Gaussian elimination chain} length. The cliff concentrates
entirely at the Recursive and Compositional levels. Full per-model,
per-subcategory accuracy including rank, nullity, trace, and transpose
breakdowns is reported in Appendix~\ref{app:accuracy} (Tables~C.1--C.3).

\begin{figure}[H]
    \centering
    \includegraphics[width=\linewidth]{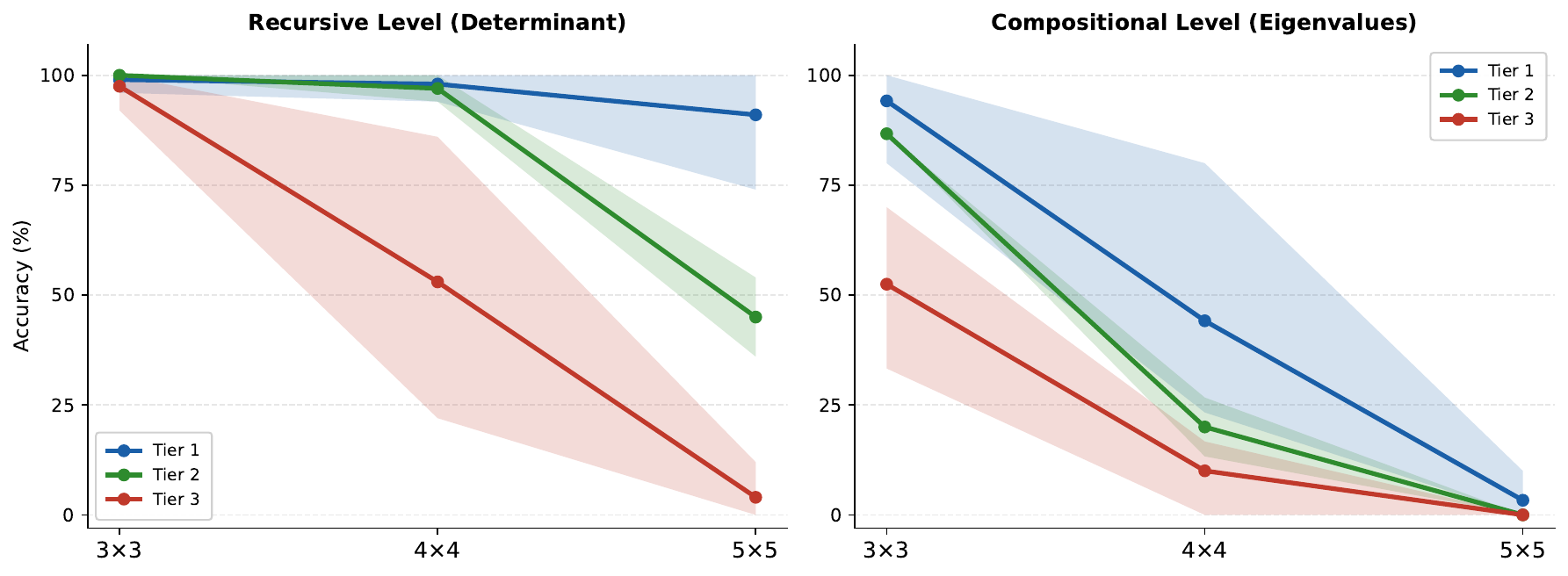}
    \caption{Accuracy trajectories across matrix dimensions for the
    Recursive level (determinant, left) and Compositional level
    (eigenvalues, right), grouped by tier. Lines show tier mean accuracy;
    shaded bands show the min--max range within each tier. Tier~1 (blue):
    OpenAI-o1, Gemini-3.0-Pro, DeepSeek-V3, Qwen3-235B. Tier~2 (green):
    GPT-5.2, Mistral-Large. Tier~3 (red): Claude-4.5-Sonnet,
    Qwen2.5-72B, Llama-3.3-70B, GPT-4o. All proportions are exact counts
    from 6,600 exhaustively evaluated outputs (2,200 per dimension; 220
    per model $\times$ 10 models).}
    \label{fig:complexity_cliff}
\end{figure}

The determinant task carries the cliff signal. At $3\times3$, all models
achieve near-perfect accuracy. At $4\times4$, the cliff begins to manifest
with the first structural collapses among lower-tier models. The decisive
break occurs at $5\times5$: a 32 percentage point step-discontinuity from
$4\times4$ to $5\times5$ determinant accuracy --- far exceeding the
gradual Sequential decline --- proving the bottleneck is recursive
state-tracking, not general difficulty scaling. Eigenvalue accuracy
collapses to near-zero across all models at $5\times5$, consistent with
its design as a deliberate stress test under maximum recursive load rather
than a standard capability measurement.

Models separate into three tiers defined by their $5\times5$ determinant
boundary. Tier~1 (OpenAI-o1, Gemini-3.0-Pro, Qwen3-235B, DeepSeek-V3)
maintains $\geq74$\% accuracy. Tier~2 (GPT-5.2, Mistral-Large) falls to
36--54\%. Tier~3 (Claude-4.5-Sonnet, Qwen2.5-72B, Llama-3.3-70B,
GPT-4o) collapses to 0--12\%. Critically, all three tiers converge to
near-zero eigenvalue accuracy by $5\times5$ --- the Recursive and
Compositional levels represent two distinct failure thresholds, not a
single continuum. This universal collapse at the Compositional level
provides stark empirical evidence for the compositionality gap
\citep{press2022compositionality}, where models fail on compound tasks
despite mastering the underlying sub-tasks.

Format sensitivity is negligible at $4\times4$ but emerges at $5\times5$.
The effect distinguishes parsing limits from computational limits: while
Arithmetic tasks (e.g., matrix-vector) maintain near-perfect accuracy
under standard LaTeX notation, altering the input to Tabular or List
formats induces artificial parsing failures --- GPT-4o drops to 50--54\%
and Claude-4.5-Sonnet to 59--61\% under non-LaTeX formats at $5\times5$,
compared to 63\% and 66\% under LaTeX respectively. At the Recursive
level, however, format preferences diverge, and fully collapsed models
fail regardless of notation --- confirming that the determinant collapse
is a fundamental working memory bottleneck, not a parsing artifact. Full
format sensitivity data is reported in Appendix~\ref{app:format}.

\section{Forensic Error Analysis}
\label{sec:forensic}

\subsection{Classification Methodology}
\label{sec:methodology}

Forensic error classification was applied to all 1,156 incorrect
responses (100 at $3\times3$, 394 at $4\times4$, 662 at $5\times5$),
following the first-error principle: the earliest step at which
computation diverges from ground truth is tagged, not the final
observable symptom. The taxonomy comprises ten primary tags applicable
across all nine subcategories --- execution errors
(\textsc{sign\_error}, \textsc{arithmetic}, \textsc{carry\_down\_error},
\textsc{memory\_loss}), structural errors (\textsc{hallucination},
\textsc{method\_fail}), and artifacts (\textsc{input\_transcription},
\textsc{generation\_truncation}, \textsc{formatting\_mismatch},
\textsc{other\_unmapped}), defined in Table~\ref{tab:error_taxonomy} ---
plus four eigenvalue-specific extensions (\textsc{generation\_loop},
\textsc{algebraic\_precedence}, \textsc{false\_verification},
\textsc{variable\_entanglement}) that arise exclusively during
characteristic polynomial expansion and root-finding, defined in
Appendix~\ref{app:judge}, Table~E.2. All distributional analysis in
Section~5 uses the ten primary tags; eigenvalue extensions are reported
in Appendix~\ref{app:error_dist}.

\begin{table}[t]
\caption{Ten primary error tags of the LinAlg-Bench forensic taxonomy,
grouped by family. Classification follows the first-error principle: the
earliest step at which computation diverges from ground truth is tagged.
Two prompt-level rules precede semantic judgment: the Magnitude Rule
($|$wrong$|$ $=$ $|$correct$|$ $\rightarrow$ \textsc{sign\_error}) and
the Truncation Precheck (no final answer $\rightarrow$
\textsc{generation\_truncation}).}
\label{tab:error_taxonomy}
\centering
\begin{tabular}{lll}
\toprule
\textbf{Tag} & \textbf{Family} & \textbf{Definition} \\
\midrule
\textsc{sign\_error}             & Execution   & Correct magnitude, wrong sign --- parity or product error \\
\textsc{arithmetic}              & Execution   & Wrong magnitude --- numerical computation error \\
\textsc{carry\_down\_error}      & Execution   & Correct at step N, miscopied at step N+1 \\
\textsc{memory\_loss}            & Execution   & Correctly stated at step N, recalled wrong at step M ($M \gg N$) \\
\textsc{hallucination}           & Structural  & Abandons or fabricates computation \\
\textsc{method\_fail}            & Structural  & Wrong algorithm applied from the first step \\
\textsc{input\_transcription}    & Artifact    & Matrix misread at input before any arithmetic \\
\textsc{generation\_truncation}  & Artifact    & Response ends without a final answer \\
\textsc{formatting\_mismatch}    & Artifact    & All values correct, wrong output format \\
\textsc{other\_unmapped}         & Residual    & Does not fit any above tag \\
\bottomrule
\end{tabular}
\end{table}

Two rules are enforced before any semantic judgment: the Magnitude Rule
($|$wrong$|$ $=$ $|$correct$|$ $\rightarrow$ \textsc{sign\_error}, never
\textsc{arithmetic}) and the Truncation Precheck (no final answer
$\rightarrow$ \textsc{generation\_truncation}, never
\textsc{hallucination}). Classification was performed by a three-stage
automated pipeline following the LLM-as-a-Judge paradigm
\citep{zheng2023judging}: the Build Judge identifies the first
divergence, the Validate Judge re-examines under adversarial framing,
and the Meta-Auditor resolves batch-level disagreements. To validate
pipeline reliability, 593 responses were independently hand-labeled by
the authors, weighted toward $5\times5$ cases where automated agreement
is lowest. Pipeline agreement with human labels reaches 89.7\% at
$5\times5$ and 92.6\% overall (Appendix~\ref{app:judge}, Table~E.4).
The five tags accounting for 91.5\% of failures achieve consistent
agreement across all dimensions; seven low-frequency tags fall below the
70\% threshold at one or more dimensions and are excluded from
distributional analysis, collectively representing only 3.6\% of all
classified failures. All 6,600 outputs were evaluated exactly once at
temperature 0; because this constitutes a complete enumeration of the
defined problem set rather than a sample, standard errors and confidence
intervals over model accuracy are not applicable.

\subsection{Failure Mode Shift}
\label{sec:failure_shift}

Failure modes shift systematically with matrix dimension.
\textsc{sign\_error} leads at $3\times3$ (33.0\%) but drops to 17.2\%
at $5\times5$. Conversely, \textsc{hallucination} rises from 17.0\% to
47.1\% --- a transition from execution failure to computational
abandonment.

\begin{figure}[H]
    \centering
    \includegraphics[width=\linewidth]{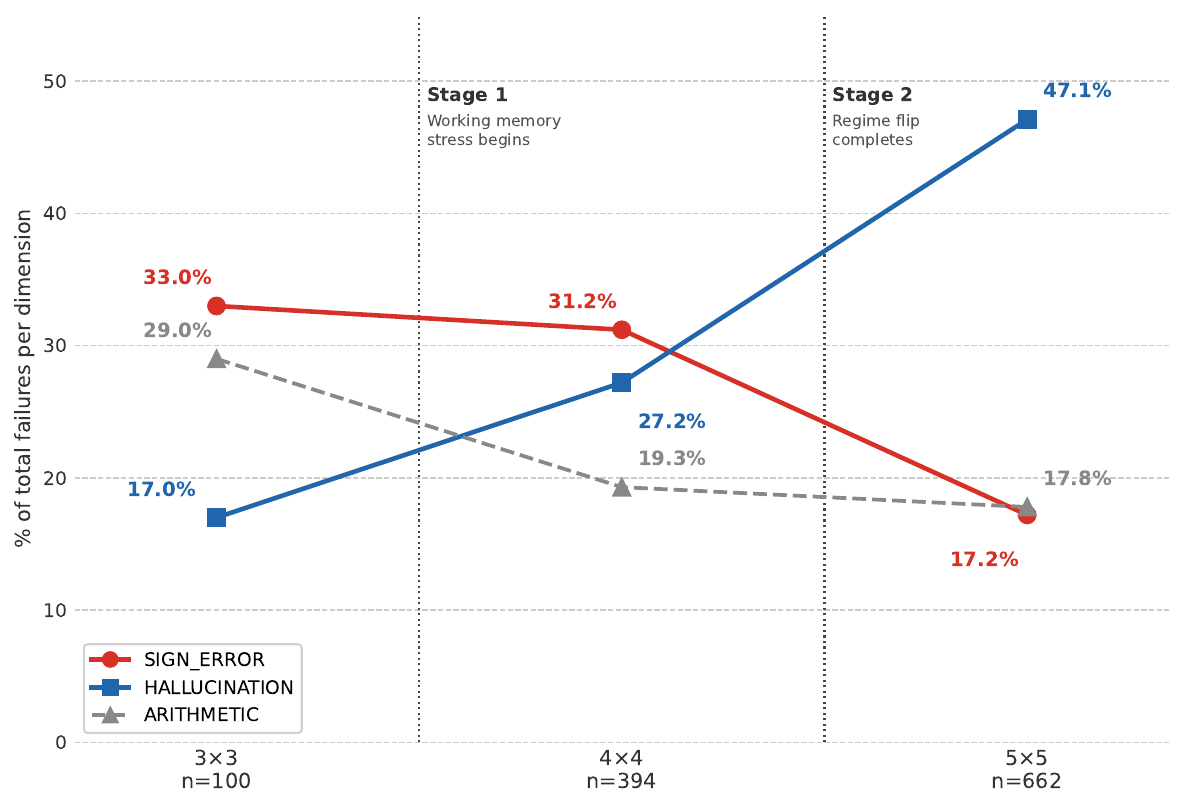}
    \caption{Error tag distribution across matrix dimensions, expressed
    as percentage of total failures per dimension ($n{=}100$ at
    $3\times3$; $n{=}394$ at $4\times4$; $n{=}662$ at $5\times5$).
    Stage~1 ($3\times3 \to 4\times4$): scale-emergent errors first appear,
    marking the onset of working memory stress. Stage~2
    ($4\times4 \to 5\times5$): the full regime flip completes ---
    \textsc{hallucination} overtakes \textsc{sign\_error} as the dominant
    failure mode.}
    \label{fig:error_regime}
\end{figure}

This collapse unfolds in two distinct stages. First, scale-emergent
errors (\textsc{generation\_loop}, \textsc{memory\_loss},
\textsc{variable\_entanglement}) appear at $4\times4$, marking the onset
of working memory stress as execution falters. Second, the full regime
flip completes at $5\times5$, where \textsc{hallucination} overtakes
\textsc{sign\_error}, driven overwhelmingly by Complete\_Collapse
(explicit abandonment). Parity errors mirror this trajectory: surging at
$4\times4$ as recursive depth stresses tracking, then retreating at
$5\times5$ as models abandon the cofactor algorithm altogether
(Appendix~\ref{app:sign}). The dominance of sign errors at lower
dimensions and their persistence across model tiers is consistent with
findings that transformers systematically struggle with signed arithmetic
operations \citep{lee2023teaching}. Sign errors are not uniformly
distributed across model tiers: Gemini-3.0-Pro contributes zero sign
errors across all dimensions, while Tier~3 models account for 221 of 270
total sign errors (81.9\%), consistent with the tier separation observed
at the Recursive level.

\subsection{Constraint-Aware Confabulation and Tool Collapse}
\label{sec:confabulation}

\textsc{hallucination} at $5\times5$ is not random. When models cannot
compute eigenvalues, fabricated responses are structured by partial
mathematical knowledge: eigenvalue sums match the matrix trace in 45\%
of fabrications, and magnitudes respect the Frobenius norm bound in 85\%
of cases ($n{=}20$ Ungrounded\_Guess cases). A benchmark evaluating only
necessary conditions would systematically misclassify these fabrications
as correct.

Across 205 Complete\_Collapse instances at $5\times5$ eigenvalue, the
failure follows a consistent three-step pattern: the model correctly
identifies the task and sets up the characteristic equation, generates a
meta-statement citing complexity, then fabricates eigenvalues via
simulated tool invocation --- citing Python, NumPy, or WolframAlpha it
does not have access to. Tool use was disabled for all models;
Gemini-3.0-Pro and Llama-3.3-70B explicitly output the system constraint
before violating it, confirming tool invocation is hallucinated rather
than real. This contrasts sharply with models explicitly trained to call
external APIs \citep{schick2023toolformer}; the tool-use syntax present
in pretraining data appears to be triggered as a collapse response when
mathematical execution is overloaded. This simulated tool invocation
likely stems from the prevalence of program-aided reasoning in
pretraining mixtures \citep{gao2023pal}; when internal computation
fails, the model falls back on code-generation syntax.

The pattern has one structurally significant exception. DeepSeek-V3
achieves 10\% eigenvalue accuracy at $5\times5$ (3/30) --- the only
open-weight model to pass any $5\times5$ eigenvalue problems --- while
using cofactor expansion 78\% of the time, a strategy profile
indistinguishable from failing models. Forensic inspection of its 3
successful responses confirms genuine computation: DeepSeek correctly
expands the full $5\times5$ characteristic polynomial before applying
bisection root-finding, independently arriving at the same numerical
strategy as OpenAI-o1. Its 27 remaining eigenvalue failures follow the
identical tool-roleplay collapse pattern --- establishing that superior
execution raises the working memory ceiling rather than eliminating it.
DeepSeek's success is not a capability advantage in any general sense;
it is a precision advantage at the specific computational boundary where
other models abandon.

Full model-specific collapse personas are documented in
Appendix~\ref{app:personas}.

\section{Interpretation}
\label{sec:interpretation}

\subsection{The Working Memory Account}
\label{sec:working_memory}

The dimensional shift documented in Section~\ref{sec:forensic} ---
execution errors giving way to computational abandonment --- is a
threshold transition, not gradual degradation. We propose a working
memory account as the most parsimonious explanation, supported by three
independent lines of evidence.

The first is the dimensional gradient. Cofactor expansion of an
$n\times n$ determinant requires maintaining $n$ active sign states
across recursive sub-expansions simultaneously --- a demand that grows
superlinearly with dimension. The same algorithm, applied by the same
models, succeeds reliably at $3\times3$ and collapses at $5\times5$. The
failure cannot be attributed to a knowledge deficit: models correctly
describe the algorithm at $5\times5$ before failing to execute it.

The second is the error regime flip. At $3\times3$, 64\% of failures are
execution errors --- models attempt computation and fail within it. At
$5\times5$, 47.1\% of failures are \textsc{hallucination}, of which
81.7\% are Complete\_Collapse (explicit abandonment; the full
Abandonment block including Silent\_Omission and Premature\_Assertion
reaches 89.4\%). Gradual degradation would produce monotonically
increasing errors of all types; the data shows a categorical mode switch.

The third is scale-emergent error types. \textsc{generation\_loop},
\textsc{memory\_loss}, and \textsc{variable\_entanglement} are all absent
at $3\times3$ and appear only at $4\times4$ and $5\times5$ ---
mechanistically impossible when computational chains are short, emerging
precisely at the dimensional boundary where working memory load
increases. The universality of the threshold across model tiers,
architectures, and training paradigms rules out a training-data
explanation.

\textsc{sign\_error} does not follow this pattern. Sign error persistence
across dimensions suggests a tokenisation-level phenomenon --- consistent
with known transformer limitations in tracking arithmetic sign state
\citep{nogueira2021arithmetic} and known BPE tokenisation instabilities
in numerical representations \citep{brown2020language} --- a separate
mechanism requiring separate explanation. Architectural hypotheses
generated by this account are documented as falsifiable conjectures in
Appendix~\ref{app:profiles}.

\subsection{Algorithmic Efficiency vs.\ Autoregressive Complexity}
\label{sec:autoregressive}

In our zero-shot evaluations, solution strategy rigidity emerged as a
near-perfect predictor of $5\times5$ determinant accuracy: models rigidly
locked into cofactor expansion collapsed entirely, while successful models
(e.g., OpenAI-o1, Gemini-3.0-Pro) employed Gaussian-dominant strategies.
The intuitive conclusion was that operation count was the primary causal
bottleneck: cofactor expansion requires $O(n!)$ operations versus
$O(n^3)$ for Gaussian elimination, saturating working memory through
recursive parity-tracking.

To test whether this predictor represented the true causal bottleneck, we
conducted a targeted ablation study. Five cofactor-dominant models
(GPT-4o, Llama-3.3-70B, Qwen2.5-72B, Mistral-Large, and
Claude-4.5-Sonnet) were explicitly instructed to abandon cofactor
expansion and use Gaussian elimination exclusively. If total operation
count were the true bottleneck, enforcing the algorithmically efficient
$O(n^3)$ path should have recovered accuracy.

\begin{figure}[H]
    \centering
    \includegraphics[width=\linewidth]{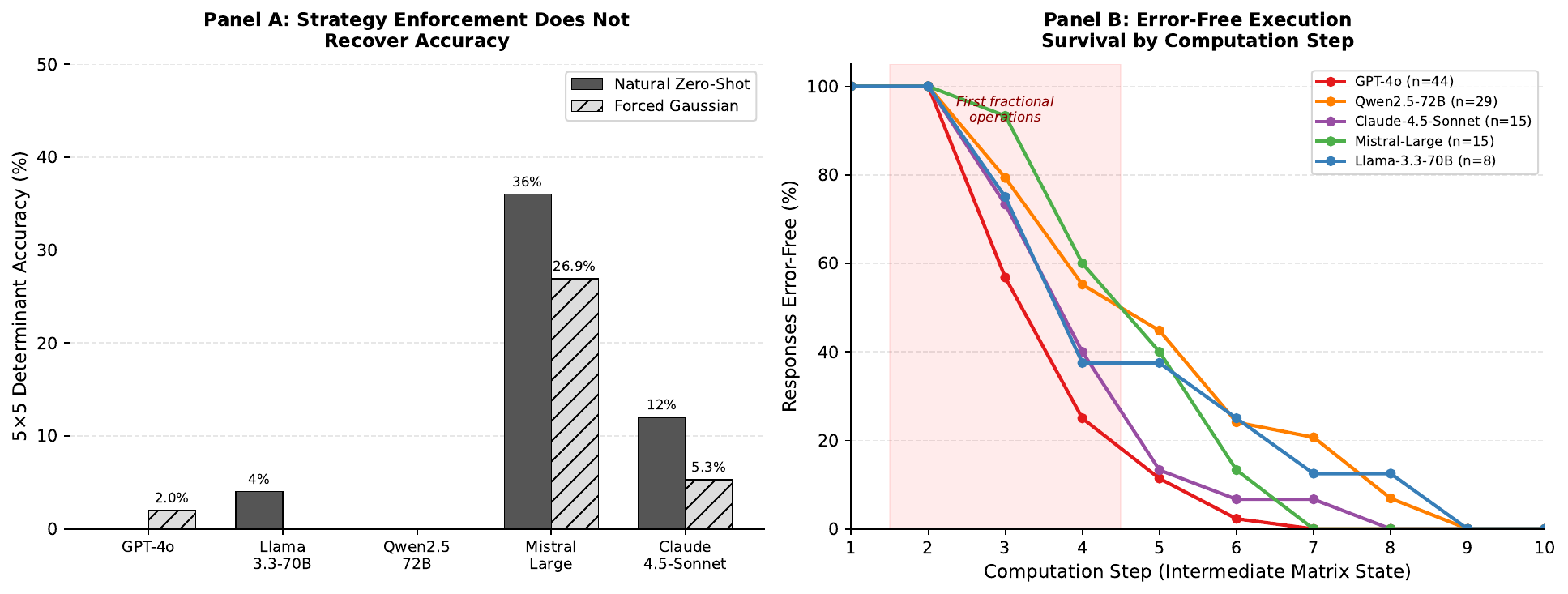}
    \caption{Forced Gaussian elimination ablation. \textbf{Panel~A}:
    $5\times5$ determinant accuracy under natural zero-shot prompting
    versus forced Gaussian elimination for five cofactor-dominant models.
    Enforcing the algorithmically efficient $O(n^3)$ strategy yields no
    meaningful accuracy recovery. Mistral-Large shows partial improvement
    (26.9\% on previously failed problems); Claude-4.5-Sonnet: 5.3\%
    ($n{=}19$). \textbf{Panel~B}: Percentage of
    responses maintaining error-free execution at each computation step.
    All models sustain error-free execution through step~2; collapse
    concentrates at steps~3--4 when the first sequentially dependent
    fractional row operations are introduced.}
    \label{fig:ablation}
\end{figure}

It did not. As shown in Figure~\ref{fig:ablation} (Panel~A), accuracy
remained at or near 0\% for fully collapsed models, and did not
meaningfully recover for partially capable models --- Mistral-Large
achieved only 26.9\% on previously failed problems, and Claude-4.5-Sonnet
a mere 5.3\% ($n{=}19$). The sole exception is DeepSeek-V3, which
achieves 74\% determinant accuracy and 10\% eigenvalue accuracy at
$5\times5$ --- the only model to sustain meaningful performance at both
the Recursive and Compositional levels while remaining cofactor-dominant
(78\%). Its success stems from superior execution quality: systematic
sub-minor simplification allows it to artificially reduce the
autoregressive depth of the cofactor algorithm, representing a unique,
partial solution to the same underlying bottleneck and demonstrating that
execution quality can partially substitute for strategy adaptation.

Forensic trace analysis (Figure~\ref{fig:ablation}, Panel~B) reveals the
mechanism and exposes a fundamental dichotomy between algorithmic
complexity and autoregressive complexity. While Gaussian elimination
minimizes total operations, it is autoregressively deep: it requires a
strictly sequential chain of highly dependent fractional arithmetic
(e.g., $R_3 \leftarrow R_3 - \frac{17}{3}R_1$). Panel~B shows that the
moment this fractional dependency is introduced --- typically between
steps 2 and 4 --- execution survival plummets across all five models.

A single hallucinated numerator or denominator early in the row-reduction
acts as a poison pill. Because the transformer must attend to its own
corrupted fractions to compute each subsequent update, the error cascades
irreversibly --- consistent with snowballing hallucination dynamics
\citep{zhang2023snowball} --- collapsing the final diagonal product to
zero or a spurious value. This is not a failure of strategy selection ---
it is a failure of precision maintenance under sequential dependency.

This structural vulnerability explains why cofactor expansion dominates at
$4\times4$ across most models --- and why Tier~3 models that remain locked
into cofactor at $5\times5$ collapse entirely. While cofactor requires
vastly more absolute operations, its steps are autoregressively flat:
broad, parallel, integer-based minor calculations with no fractional
carry-forward between terms. It is mathematically safer for a transformer
to execute a high volume of independent integer steps than a low volume of
deeply dependent fractional ones. The primary bottleneck in LLM
mathematical reasoning is not the volume of computation, but the model's
inability to maintain strict numerical attention across deep, sequentially
dependent fractional chains without error propagation --- a failure mode
that process reward models were specifically designed to address
\citep{lightman2023verify}.

\section{Discussion}
\label{sec:discussion}

LinAlg-Bench contributes to evaluation methodology by demonstrating that
diagnostic benchmarks can generate falsifiable mechanistic hypotheses
rather than merely leaderboard rankings. A benchmark evaluating
determinants at a single matrix size would classify Llama-3.3-70B,
GPT-4o, and Claude-4.5-Sonnet as capable --- all achieve $\geq98$\% at
$3\times3$ --- while concealing complete collapse at $5\times5$. The
fabrication-to-abandonment threshold, the two-stage working memory
account, and the constraint-aware confabulation finding are each directly
testable by independent replication and mechanistic interpretability
tools.

Several limitations constrain the scope of claims. First, the benchmark
covers linear algebra only --- a domain with precise algorithmic
structure and exact verifiable answers. Whether the
fabrication-to-abandonment threshold generalises to other mathematical
domains remains an open empirical question. Second, all models were
evaluated at temperature 0 with a single run per problem;
higher-temperature majority voting may alter the threshold location, and
self-consistency sampling is a natural extension. Third, integer-entry
matrices and exact-match evaluation eliminate floating-point ambiguity
but exclude numerical robustness as a failure surface. Fourth, tool
availability and API-level constraints mean that some model behaviours
--- particularly GPT-5.2's apparent tool circumvention --- cannot be
definitively attributed to specific implementation choices without access
to inference infrastructure.

Three directions follow directly. First, constraint-aware confabulation
warrants cross-domain investigation: whether LLMs fabricate
constraint-consistent answers under computational overload in chemistry,
physics, or statistics is a testable hypothesis with direct evaluation
methodology implications. Second, the working memory account generates
specific mechanistic predictions testable by activation patching at the
cofactor expansion steps. Third, the ablation finding --- that enforcing
correct strategy does not recover accuracy --- suggests that multi-step
reasoning training curricula should be evaluated on their ability to
induce execution robustness, not just accuracy at fixed difficulty.

\section{Conclusion}
\label{sec:conclusion}

LinAlg-Bench reveals that LLM mathematical failure is not random but
structurally constrained --- determined by algorithm family, matrix
dimension, and autoregressive execution depth. The
fabrication-to-abandonment threshold at $4\times4$ scale, near-universal
across model tiers and architectures, is consistent with a working memory
limit rather than a knowledge gap. Two supporting findings strengthen
this account: scale-emergent error types appear at precisely the
dimensional boundary where recursive depth increases, and a
forced-strategy ablation shows that enforcing algorithmically efficient
methods does not recover accuracy --- the bottleneck is execution depth,
not method selection. Constraint-aware confabulation --- models
fabricating eigenvalues that satisfy verifiable mathematical constraints
--- demonstrates that hallucination under computational overload is
structured by partial knowledge rather than arbitrary. These are
falsifiable claims that follow directly from the data and invite
verification by mechanistic interpretability tools and cross-domain
replication. All data, model outputs, error labels, and judge pipeline
are publicly released.


\paragraph{Data Availability.} All benchmark problems, model outputs, error annotations, and results are publicly available at \url{https://huggingface.co/datasets/LinAlgBench/linalg-bench}. Code and evaluation scripts are available at \url{https://github.com/shradhautk/LinAlgBench}.
\bibliographystyle{plainnat}
\bibliography{references}

\newpage

\appendix


\nolinenumbers

\section{Benchmark Inference Prompts}
\label{app:prompts}

All models were evaluated zero-shot at temperature~0. The prompt consists of a
system prompt and a user prompt, both fixed per subcategory. A base zero-shot
chain-of-thought (ZS-CoT) instruction was included to elicit reasoning
traces~\citep{wei2022chain,kojima2022zero} (\emph{``Show all computation steps
clearly''}; \emph{``Show your work''}) but no few-shot examples, method
guidance, or algorithm-specific instructions were provided. Tool use was
disabled for all models via API configuration.

\subsection{System Prompt}
\label{app:system-prompt}

The system prompt is shared across all subcategories. Only the answer-type
directive (\underline{underlined below}) varies:

\bigskip
\noindent\fbox{%
  \parbox{0.93\linewidth}{%
    You are a precise mathematical assistant.\\[2pt]
    Show all computation steps clearly.\\[2pt]
    Always put your final \underline{[answer-type]} answer inside
    \texttt{\textbackslash boxed\{\}}.
  }%
}
\bigskip

\begin{table}[H]
  \centering
  \caption{Answer-type directive by subcategory.}
  \label{tab:answer-type}
  \small
  \begin{tabular}{lll}
    \toprule
    \textbf{Subcategory} & \textbf{Answer-type word} & \textbf{Tasks} \\
    \midrule
    \texttt{det}, \texttt{trace}               & numerical & Determinant, Trace \\
    \texttt{eig}                               & answers   & Eigenvalues \\
    \texttt{rank}, \texttt{null}               & integer   & Rank, Nullity \\
    \texttt{mult}, \texttt{pow}, \texttt{trans} & matrix   & Mult., Power, Transpose \\
    \texttt{vec}                               & vector    & Matrix-Vector \\
    \bottomrule
  \end{tabular}
\end{table}

\subsection{User Prompt Template}
\label{app:user-prompt}

The user prompt template is identical across all nine subcategories:

\bigskip
\noindent\fbox{%
  \parbox{0.93\linewidth}{%
    Solve this linear algebra problem.\\[2pt]
    Show your work and give the final answer.\\[8pt]
    \texttt{\{problem\_text\}}\\[8pt]
    \texttt{\{problem\_matrix\}}
  }%
}
\bigskip

\noindent where \texttt{\{problem\_text\}} is the task instruction (e.g.\
``Compute the determinant of matrix~$A$'') and \texttt{\{problem\_matrix\}} is
the matrix in one of three input formats described in
Section~\ref{app:formats}.

\subsection{Input Format Examples}
\label{app:formats}

The same $3{\times}3$ determinant problem is shown below in all three input
formats used in the benchmark. Format sensitivity was evaluated at $4{\times}4$
and $5{\times}5$ only; the $3{\times}3$ dimension is shown here for brevity.

\begin{table}[H]
  \centering
  \caption{Three input format representations of the same determinant problem.
           All three encode identical numerical content; only the notation
           differs.}
  \label{tab:formats}
  \small
  \renewcommand{\arraystretch}{1.4}
  \begin{tabular}{>{\bfseries}p{1.4cm} p{5.8cm} p{5.0cm}}
    \toprule
    Format & \texttt{\{problem\_matrix\}} value & Description \\
    \midrule
    LaTeX
      & $\displaystyle A = \begin{bmatrix}
           2 & -1 &  3 \\
           0 &  4 & -2 \\
          -1 &  1 &  5
        \end{bmatrix}$
      & Standard \LaTeX\ \texttt{bmatrix} environment. Provides explicit
        algebraic notation. \\[6pt]
    Tabular
      & \ttfamily\small
        \begin{tabular}[t]{@{}l@{}}
          \textbar\ \ 2\ \ -1\ \ \ 3\ \textbar \\
          \textbar\ \ 0\ \ \ 4\ \ -2\ \textbar \\
          \textbar\ -1\ \ \ 1\ \ \ 5\ \textbar
        \end{tabular}
      & Pipe-delimited human-readable grid. Visual row-column structure without
        mathematical notation. \\[6pt]
    List
      & \ttfamily\small
        \begin{tabular}[t]{@{}l@{}}
          A = [[2, -1, 3], \\
          \phantom{A = [}[0,\phantom{-}\ 4, -2], \\
          \phantom{A = [}[-1, 1,\phantom{-}\ 5]]
        \end{tabular}
      & Nested Python-style list. No explicit visual alignment. \\
    \bottomrule
  \end{tabular}
\end{table}

\subsection{Model Details and Selection Rationale}
\label{app:models}

Table~\ref{tab:model-ids} reports the exact model identifiers and access
configuration used for all benchmark inference runs, sourced directly from the
production inference code. All models were evaluated zero-shot at temperature~0
with tool use and external code execution disabled. Token budgets were set to
8{,}192 for most subcategories and 16{,}384 for DeepSeek-V3, GPT-5.2, and
Qwen3-235B. Responses reaching the token ceiling without producing a final
answer were classified as \textsc{generation\_truncation}, strictly separating
infrastructure limits from computational collapse.

\paragraph{Reproducibility window.}
To establish a reproducible temporal baseline, all inference across all 10
models was conducted within a strict window between 25~December~2025 and
3~May~2026. Six models were accessed via the OpenRouter routing
layer~\citep{openrouter2024}: DeepSeek-V3~\citep{deepseek2024v3},
Qwen3-235B and Qwen2.5-72B~\citep{qwen2025qwen3,qwen2024qwen25},
Mistral-Large~\citep{mistral2024large},
Claude-4.5-Sonnet~\citep{anthropic2024claude}, and
Llama-3.3-70B~\citep{meta2024llama3}. OpenRouter does not guarantee snapshot
pinning; the date window bounds this uncertainty. Four models were accessed via
direct provider APIs: OpenAI-o1, GPT-4o, and GPT-5.2 via the OpenAI
API~\citep{openai2024gpt4,openai2024o1}, and Gemini-3.0-Pro via the Google AI
Studio GenAI SDK~\citep{google2024gemini}.

\begin{table}[H]
  \centering
  \caption{Model identifiers and inference configuration.}
  \label{tab:model-ids}
  \footnotesize
  \renewcommand{\arraystretch}{1.15}
  \resizebox{\linewidth}{!}{%
  \begin{tabular}{lllll}
    \toprule
    \textbf{Paper Name} & \textbf{Model ID} & \textbf{Access} &
    \textbf{Provider} & \textbf{Architecture} \\
    \midrule
    OpenAI-o1         & \texttt{o1}                                & Direct     & OpenAI           & Dense, reasoning-optimised \\
    Gemini-3.0-Pro    & \texttt{gemini-3.1-pro-preview}            & Direct     & Google AI Studio & Dense, frontier \\
    GPT-5.2           & \texttt{gpt-5.2}                           & Direct     & OpenAI           & Dense, frontier \\
    GPT-4o            & \texttt{gpt-4o}                            & Direct     & OpenAI           & Dense, instruction-tuned \\
    DeepSeek-V3       & \texttt{deepseek/deepseek-v3.2}            & OpenRouter & DeepSeek         & MoE, instruction-tuned \\
    Qwen3-235B        & \texttt{qwen/qwen3-235b-a22b-2507}         & OpenRouter & Alibaba          & MoE, thinking/non-thinking hybrid \\
    Mistral-Large     & \texttt{mistralai/mistral-large-2512}      & OpenRouter & Mistral AI       & MoE, instruction-tuned \\
    Claude-4.5-Sonnet & \texttt{anthropic/claude-sonnet-4.5}       & OpenRouter & Anthropic        & Dense, instruction-tuned \\
    Qwen2.5-72B       & \texttt{qwen/qwen-2.5-72b-instruct}        & OpenRouter & Alibaba          & Dense, instruction-tuned \\
    Llama-3.3-70B     & \texttt{meta-llama/llama-3.3-70b-instruct} & OpenRouter & Meta             & Dense, instruction-tuned \\
    \bottomrule
  \end{tabular}}
  \vspace{4pt}
  \\\footnotesize\textit{Note.} All model IDs copied verbatim from production
  inference code. Paper name Gemini-3.0-Pro refers to model ID
  \texttt{gemini-3.1-pro-preview}. Paper name DeepSeek-V3 refers to the
  base/chat MoE model \texttt{deepseek-v3.2}; this is not the DeepSeek-R1
  reasoning model.
\end{table}

\subsection{Model Selection Rationale}
\label{app:selection}

The 10 models were selected to cover five independent axes of variation,
enabling the benchmark to test whether failure patterns are specific to
particular training regimes, architectures, or providers.

\begin{table}[H]
  \centering
  \caption{Selection rationale. Models were selected to cover five independent
           axes of variation: training paradigm, architecture, provider, weight
           availability, and scale. No model was selected on the basis of
           expected performance outcome. Selection was finalised before any
           benchmark runs were conducted.}
  \label{tab:selection}
  \small
  \renewcommand{\arraystretch}{1.3}
  \begin{tabular}{>{\bfseries}p{2.8cm} p{10.5cm}}
    \toprule
    Axis & Coverage \\
    \midrule
    Training paradigm
      & Reasoning-optimised (OpenAI-o1); thinking-mode hybrid (Qwen3-235B);
        standard RLHF dense (Llama-3.3-70B, Qwen2.5-72B, GPT-4o,
        Claude-4.5-Sonnet); balanced frontier (GPT-5.2, Gemini-3.0-Pro);
        MoE instruction-tuned (DeepSeek-V3, Mistral-Large). \\
    Architecture
      & Dense: GPT-4o, GPT-5.2, OpenAI-o1, Claude-4.5-Sonnet, Qwen2.5-72B,
        Llama-3.3-70B, Gemini-3.0-Pro.
        MoE: Qwen3-235B (22B active / 235B total), Mistral-Large,
        DeepSeek-V3. \\
    Provider
      & OpenAI~(3); Google~(1); Anthropic~(1); Meta~(1);
        Alibaba~(2); Mistral AI~(1); DeepSeek~(1). \\
    Weight availability
      & Closed-weight: GPT-4o, GPT-5.2, OpenAI-o1, Gemini-3.0-Pro,
        Claude-4.5-Sonnet.\quad
        Open-weight: DeepSeek-V3, Qwen3-235B, Qwen2.5-72B, Llama-3.3-70B,
        Mistral-Large. \\
    Scale
      & Active parameter range from 70B (Llama-3.3-70B, Qwen2.5-72B) to
        235B total / 22B active (Qwen3-235B). Closed-weight models not
        directly comparable by parameter count. \\
    \bottomrule
  \end{tabular}
\end{table}

\subsection{Excluded Models}
\label{app:excluded}

Qwen-2.5-Math-72B was accessed via a HuggingFace Space inference
endpoint---not an API-controlled environment---and is excluded as results are
not reproducible under the same inference protocol. Claude-3.5-Sonnet was
registered as a standby model and replaced by Claude-4.5-Sonnet before
benchmark runs began; no Claude-3.5-Sonnet results are included in the
benchmark.


\section{Benchmark Task Descriptions}
\label{app:tasks}

This appendix describes all nine subcategories in the LinAlg-Bench benchmark.
For each task we show the exact instruction seen by the model, the algorithmic
steps required for a correct solution, and how computational demand scales
across the three matrix dimensions. Tasks are grouped by cognitive level.


\paragraph{Trace \normalfont\textit{· Cognitive level: Reading}}
\mbox{}\\[2pt]
\noindent\textbf{Instruction:} Compute the trace of matrix $A$. Put your final
numerical answer inside $\backslash$\texttt{boxed\{\}}.

\noindent\textbf{Required steps:} Sum the main diagonal entries
$a_{11} + a_{22} + \cdots + a_{nn}$.

\noindent\textbf{Expected output:} Single integer scalar.

\begin{center}
\small
\begin{tabular}{lccc}
  \toprule
                    & \textbf{3$\times$3} & \textbf{4$\times$4} & \textbf{5$\times$5} \\
  \midrule
  Operations        & 2 additions         & 3 additions         & 4 additions         \\
  \bottomrule
\end{tabular}
\end{center}

\noindent\textbf{Primary failure at 5$\times$5:} \textit{None --- trace is
format-invariant and dimension-invariant. Serves as ceiling baseline.}

\bigskip


\paragraph{Transpose \normalfont\textit{· Cognitive level: Reading}}
\mbox{}\\[2pt]
\noindent\textbf{Instruction:} Compute the transpose of matrix $A$. Put your
final matrix answer inside $\backslash$\texttt{boxed\{\}}.

\noindent\textbf{Required steps:} Reflect all entries across the main diagonal:
entry $(i,j)$ moves to $(j,i)$.

\noindent\textbf{Expected output:} $n \times n$ matrix.

\begin{center}
\small
\begin{tabular}{lccc}
  \toprule
                     & \textbf{3$\times$3}      & \textbf{4$\times$4}      & \textbf{5$\times$5}       \\
  \midrule
  Swaps required     & 3 off-diagonal pairs     & 6 off-diagonal pairs     & 10 off-diagonal pairs     \\
  \bottomrule
\end{tabular}
\end{center}

\noindent\textbf{Primary failure at 5$\times$5:} \textit{None --- purely
positional, no arithmetic. Serves as ceiling baseline.}

\bigskip


\paragraph{Matrix-Vector Multiplication \normalfont\textit{· Cognitive level: Arithmetic}}
\mbox{}\\[2pt]
\noindent\textbf{Instruction:} Compute the matrix-vector product $Ax$. Put your
final vector answer inside $\backslash$\texttt{boxed\{\}}.

\noindent\textbf{Required steps:} For each row of $A$, compute the dot product
with vector $x$: sum of $a_{ij} \cdot x_j$ across $j$.

\noindent\textbf{Expected output:} Column vector of length $n$.

\begin{center}
\small
\begin{tabular}{lccc}
  \toprule
                   & \textbf{3$\times$3}  & \textbf{4$\times$4}  & \textbf{5$\times$5}  \\
  \midrule
  Dot products     & 3 (one per row)      & 4 (one per row)      & 5 (one per row)      \\
  Multiplications  & 9 total              & 16 total             & 25 total             \\
  Sign risk        & Low                  & Moderate             & Moderate             \\
  \bottomrule
\end{tabular}
\end{center}

\noindent\textbf{Primary failure at 5$\times$5:} \textit{Product\_Sign\_Error
and Double\_Negative\_Trap when operands are negative.}

\bigskip


\paragraph{Matrix Multiplication \normalfont\textit{· Cognitive level: Arithmetic}}
\mbox{}\\[2pt]
\noindent\textbf{Instruction:} Compute the matrix product $AB$, where $A$ and
$B$ are both $n \times n$ matrices. Put your final matrix answer inside
$\backslash$\texttt{boxed\{\}}.

\noindent\textbf{Required steps:} For each output entry $(i,j)$, compute dot
product of row $i$ of $A$ with column $j$ of $B$.

\noindent\textbf{Expected output:} $n \times n$ matrix.

\begin{center}
\small
\begin{tabular}{lccc}
  \toprule
                   & \textbf{3$\times$3}         & \textbf{4$\times$4}          & \textbf{5$\times$5}          \\
  \midrule
  Output entries   & 9                           & 16                           & 25                           \\
  Dot products     & 9 (length 3 each)           & 16 (length 4 each)           & 25 (length 5 each)           \\
  Multiplications  & 27 total                    & 64 total                     & 125 total                    \\
  \bottomrule
\end{tabular}
\end{center}

\noindent\textbf{Primary failure at 5$\times$5:} \textit{Rule\_Interference and
Double\_Negative\_Trap; errors compound across all 25 entries.}

\bigskip


\paragraph{Matrix Power \normalfont\textit{· Cognitive level: Arithmetic}}
\mbox{}\\[2pt]
\noindent\textbf{Instruction:} Compute $A^k$. Put your final matrix answer
inside $\backslash$\texttt{boxed\{\}}.

\noindent\textbf{Required steps:} Compute $A^2 = A \cdot A$, then
$A^3 = A^2 \cdot A$, repeating $k-1$ multiplications. Each multiplication
requires $n^2$ dot products of length $n$.

\noindent\textbf{Expected output:} $n \times n$ matrix.

\begin{center}
\small
\renewcommand{\arraystretch}{1.3}
\resizebox{\linewidth}{!}{\begin{tabular}{>{\raggedright\arraybackslash}p{2.6cm}
                >{\raggedright\arraybackslash}p{3.2cm}
                >{\raggedright\arraybackslash}p{3.2cm}
                >{\raggedright\arraybackslash}p{3.8cm}}
  \toprule
                     & \textbf{3$\times$3}                                  & \textbf{4$\times$4}                                  & \textbf{5$\times$5}                                        \\
  \midrule
  Multiplications    & $k-1$ full matrix multiplications                    & $k-1$ full matrix multiplications                    & $k-1$ full matrix multiplications                          \\
  Error propagation  & Sign errors from step $t$ affect all $t{+}1, t{+}2, \ldots$ & Sign errors from step $t$ affect all $t{+}1, t{+}2, \ldots$ & Sign errors compound most severely at 5$\times$5 (125 ops per step) \\
  \bottomrule
\end{tabular}}
\end{center}

\noindent\textbf{Primary failure at 5$\times$5:} \textit{Compounding sign
errors --- a single Product\_Sign\_Error in an early power corrupts all
subsequent powers.}

\bigskip


\paragraph{Rank \normalfont\textit{· Cognitive level: Sequential}}
\mbox{}\\[2pt]
\noindent\textbf{Instruction:} Compute the rank of matrix $A$. Put your final
integer answer inside $\backslash$\texttt{boxed\{\}}.

\noindent\textbf{Required steps:} Perform row reduction (Gaussian elimination)
to echelon form. Count non-zero pivot rows.

\noindent\textbf{Expected output:} Integer in $\{0, 1, \ldots, n\}$.

\begin{center}
\small
\renewcommand{\arraystretch}{1.3}
\resizebox{\linewidth}{!}{\begin{tabular}{>{\raggedright\arraybackslash}p{2.6cm}
                >{\raggedright\arraybackslash}p{3.2cm}
                >{\raggedright\arraybackslash}p{3.2cm}
                >{\raggedright\arraybackslash}p{3.8cm}}
  \toprule
                  & \textbf{3$\times$3}                       & \textbf{4$\times$4}                       & \textbf{5$\times$5}                        \\
  \midrule
  Row operations  & Up to 3 elimination steps                 & Up to 6 elimination steps                 & Up to 10 elimination steps                 \\
  Pivot decisions & 3 (is row zero?)                          & 4                                         & 5                                          \\
  Arithmetic ops  & ${\sim}9$ multiplications $+$ subtractions & ${\sim}24$ multiplications $+$ subtractions & ${\sim}50$ multiplications $+$ subtractions \\
  \bottomrule
\end{tabular}}
\end{center}

\noindent\textbf{Primary failure at 5$\times$5:} \textit{Memory\_Loss ---
model loses earlier pivot state when deciding whether late rows are zero.}

\bigskip


\paragraph{Nullity \normalfont\textit{· Cognitive level: Sequential}}
\mbox{}\\[2pt]
\noindent\textbf{Instruction:} Compute the nullity of matrix $A$. Put your
final integer answer inside $\backslash$\texttt{boxed\{\}}.

\noindent\textbf{Required steps:} Compute rank via row reduction, then apply
rank-nullity theorem: nullity $= n - \text{rank}$.

\noindent\textbf{Expected output:} Integer in $\{0, 1, \ldots, n\}$.

\begin{center}
\small
\resizebox{\linewidth}{!}{\begin{tabular}{lccc}
  \toprule
                & \textbf{3$\times$3}              & \textbf{4$\times$4}              & \textbf{5$\times$5}              \\
  \midrule
  Depends on    & Rank computation (same steps)    & Rank computation (same steps)    & Rank computation (same steps)    \\
  Extra step    & One subtraction: $3 - \text{rank}$ & One subtraction: $4 - \text{rank}$ & One subtraction: $5 - \text{rank}$ \\
  \bottomrule
\end{tabular}}
\end{center}

\noindent\textbf{Primary failure at 5$\times$5:} \textit{Inherits all rank
failures; additionally, off-by-one errors in pivot counting propagate
directly.}

\bigskip


\paragraph{Determinant \normalfont\textit{· Cognitive level: Recursive}}
\mbox{}\\[2pt]
\noindent\textbf{Instruction:} Compute the determinant of matrix $A$. Put your
final numerical answer inside $\backslash$\texttt{boxed\{\}}.

\noindent\textbf{Required steps:} Cofactor expansion:
$\det(A) = \sum_j a_{1j} \cdot (-1)^{1+j} \cdot M_{1j}$
where $M_{1j}$ is the $(n-1)\times(n-1)$ minor. Alternatively: Gaussian
elimination to upper triangular form, then multiply diagonal.

\noindent\textbf{Expected output:} Single integer scalar.

\begin{center}
\small
\renewcommand{\arraystretch}{1.3}
\resizebox{\linewidth}{!}{\begin{tabular}{>{\raggedright\arraybackslash}p{2.6cm}
                >{\raggedright\arraybackslash}p{3.2cm}
                >{\raggedright\arraybackslash}p{3.2cm}
                >{\raggedright\arraybackslash}p{3.8cm}}
  \toprule
                    & \textbf{3$\times$3}             & \textbf{4$\times$4}                  & \textbf{5$\times$5}                   \\
  \midrule
  Cofactor terms    & 3 sub-minors                    & $4 \times 3 = 12$ sub-minors         & $5 \times 4 \times 3 = 60$ sub-minors \\
  Parity checks     & 3\ \ $(+{-}+)$                  & 12\ \ (two levels of $+{-}+{-}$)     & 60+\ \ (three nested levels)          \\
  Gaussian steps    & 3 pivot eliminations            & 6 pivot eliminations                 & 10 pivot eliminations                 \\
  Method advantage  & Equal --- both manageable       & Gaussian ${\sim}3\times$ fewer ops   & Gaussian ${\sim}6\times$ fewer ops    \\
  \bottomrule
\end{tabular}}
\end{center}

\noindent\textbf{Primary failure at 5$\times$5:} \textit{Alternating\_Drift
(cofactor models lose $\pm1$ checkerboard after 3--4 terms);
Gaussian-dominant models maintain high accuracy. Strategy is the primary
predictor of success at 5$\times$5.}

\bigskip


\paragraph{Eigenvalues \normalfont\textit{· Cognitive level: Compositional}}
\mbox{}\\[2pt]
\noindent\textbf{Instruction:} Find all eigenvalues of matrix $A$. Put your
final answers inside $\backslash$\texttt{boxed\{\}}.

\noindent\textbf{Required steps:} Stage~1: Form $A - \lambda I$. Stage~2:
Expand $\det(A - \lambda I)$ to obtain characteristic polynomial of degree $n$.
Stage~3: Find all roots of the polynomial.

\noindent\textbf{Expected output:} Set of $n$ eigenvalues (may be real or
complex).

\begin{center}
\small
\renewcommand{\arraystretch}{1.3}
\resizebox{\linewidth}{!}{\begin{tabular}{>{\raggedright\arraybackslash}p{2.6cm}
                >{\raggedright\arraybackslash}p{3.2cm}
                >{\raggedright\arraybackslash}p{3.2cm}
                >{\raggedright\arraybackslash}p{3.8cm}}
  \toprule
                       & \textbf{3$\times$3}                    & \textbf{4$\times$4}                             & \textbf{5$\times$5}                                            \\
  \midrule
  Characteristic poly  & Degree 3 (cubic)                       & Degree 4 (quartic)                              & Degree 5 (quintic)                                             \\
  Cofactor terms       & 3 (each contains $\lambda$)            & 12 (each a polynomial in $\lambda$)             & 60 (each a polynomial in $\lambda$)                            \\
  Root finding         & Factorable by hand                     & Difficult, may require numerical methods        & Factorable using rational root theorem; benchmark problems have exact integer roots \\
  Failure surfaces     & 2 (det expansion $+$ root finding)     & 2 (compounded)                                  & 2 (fully compounded --- det load $\times$ polynomial load)     \\
  \bottomrule
\end{tabular}}
\end{center}

\noindent\textbf{Primary failure at 5$\times$5:} \textit{Complete\_Collapse
dominates at 5$\times$5 (80.4\% of failures): models correctly identify the
task, then abandon computation with meta-statements citing complexity,
fabricating eigenvalues via simulated tool calls.
\textbf{DeepSeek-V3} and \textbf{OpenAI-o1} demonstrate the task is solvable
with sufficient execution capacity.}


\section{Full Accuracy Tables}
\label{app:accuracy}

Tables~\ref{tab:c1}--\ref{tab:c3} report complete per-model, per-task accuracy
across all three matrix dimensions. All models were evaluated zero-shot at
temperature~0. The benchmark comprises 9 tasks across 5 cognitive levels, with
220 questions per dimension (660 total). Ground truth verified using SymPy
symbolic computation~\citep{meurer2017sympy}. Models ordered by tier
membership.

\begin{table}[p]
  \centering
  \caption{Complete accuracy table --- 3$\times$3 matrices (220 questions, 10 models). Numbers show correct responses out of maximum per category.}
  \label{tab:c1}
  \renewcommand{\arraystretch}{1.15}
  \resizebox{\linewidth}{!}{%
  \begin{tabular}{ll r rrrrrrrrrr}
  \toprule
  \textbf{Level} & \textbf{Category} & \textbf{Max} &
  \textbf{OpenAI-o1} &
  \textbf{Gemini-3.0-Pro} &
  \textbf{DeepSeek-V3} &
  \textbf{GPT-5.2} &
  \textbf{Qwen3-235B} &
  \textbf{Mistral-Large} &
  \textbf{Claude-4.5-Son.} &
  \textbf{Qwen2.5-72B} &
  \textbf{Llama-3.3-70B} &
  \textbf{GPT-4o} \\
  \midrule
  \multirow{2}{*}{\textbf{Reading}} & Trace & 20 & 20 & 20 & 20 & 20 & 20 & 20 & 20 & 20 & 20 & 20 \\
   & Transpose & 20 & 20 & 20 & 20 & 20 & 20 & 20 & 20 & 19 & 20 & 20 \\
  \multicolumn{3}{l}{\textit{\textbf{Level Accuracy (\%)}}} & \textbf{100.0} & \textbf{100.0} & \textbf{100.0} & \textbf{100.0} & \textbf{100.0} & \textbf{100.0} & \textbf{100.0} & \textbf{97.5} & \textbf{100.0} & \textbf{100.0} \\
  \addlinespace[2pt]
  \multirow{3}{*}{\textbf{Arithmetic}} & Matrix-Vector & 20 & 20 & 20 & 20 & 20 & 20 & 20 & 20 & 20 & 20 & 20 \\
   & Multiplication & 20 & 20 & 20 & 20 & 20 & 20 & 20 & 20 & 20 & 20 & 19 \\
   & Matrix Power & 20 & 20 & 20 & 20 & 20 & 13 & 20 & 20 & 20 & 19 & 20 \\
  \multicolumn{3}{l}{\textit{\textbf{Level Accuracy (\%)}}} & \textbf{100.0} & \textbf{100.0} & \textbf{100.0} & \textbf{100.0} & \textbf{88.3} & \textbf{100.0} & \textbf{100.0} & \textbf{100.0} & \textbf{98.3} & \textbf{98.3} \\
  \addlinespace[2pt]
  \multirow{2}{*}{\textbf{Sequential}} & Nullity & 20 & 20 & 20 & 20 & 20 & 20 & 20 & 20 & 19 & 14 & 20 \\
   & Rank & 20 & 20 & 20 & 20 & 20 & 19 & 20 & 20 & 19 & 19 & 19 \\
  \multicolumn{3}{l}{\textit{\textbf{Level Accuracy (\%)}}} & \textbf{100.0} & \textbf{100.0} & \textbf{100.0} & \textbf{100.0} & \textbf{97.5} & \textbf{100.0} & \textbf{100.0} & \textbf{95.0} & \textbf{82.5} & \textbf{97.5} \\
  \addlinespace[2pt]
  \multirow{1}{*}{\textbf{Deep Recursive}} & Determinant & 50 & 50 & 50 & 50 & 50 & 48 & 50 & 50 & 46 & 50 & 49 \\
  \multicolumn{3}{l}{\textit{\textbf{Level Accuracy (\%)}}} & \textbf{100.0} & \textbf{100.0} & \textbf{100.0} & \textbf{100.0} & \textbf{96.0} & \textbf{100.0} & \textbf{100.0} & \textbf{92.0} & \textbf{100.0} & \textbf{98.0} \\
  \addlinespace[2pt]
  \multirow{1}{*}{\textbf{Built on Det.}} & Eigenvalues & 30 & 30 & 30 & 29 & 26 & 24 & 26 & 21 & 19 & 10 & 13 \\
  \multicolumn{3}{l}{\textit{\textbf{Level Accuracy (\%)}}} & \textbf{100.0} & \textbf{100.0} & \textbf{96.7} & \textbf{86.7} & \textbf{80.0} & \textbf{86.7} & \textbf{70.0} & \textbf{63.3} & \textbf{33.3} & \textbf{43.3} \\
  \addlinespace[2pt]
  \midrule
  \textbf{Total Correct} & & \textbf{220} & \textbf{220} & \textbf{220} & \textbf{219} & \textbf{216} & \textbf{204} & \textbf{216} & \textbf{211} & \textbf{202} & \textbf{192} & \textbf{200} \\
  \multicolumn{3}{l}{\textbf{Overall Model Accuracy (\%)}} & \textbf{100.0} & \textbf{100.0} & \textbf{99.5} & \textbf{98.2} & \textbf{92.7} & \textbf{98.2} & \textbf{95.9} & \textbf{91.8} & \textbf{87.3} & \textbf{90.9} \\
  \bottomrule
  \end{tabular}}
\end{table}

\begin{table}[p]
  \centering
  \caption{Complete accuracy table --- 4$\times$4 matrices (220 questions, 10 models).}
  \label{tab:c2}
  \renewcommand{\arraystretch}{1.15}
  \resizebox{\linewidth}{!}{%
  \begin{tabular}{ll r rrrrrrrrrr}
  \toprule
  \textbf{Level} & \textbf{Category} & \textbf{Max} &
  \textbf{OpenAI-o1} &
  \textbf{Gemini-3.0-Pro} &
  \textbf{DeepSeek-V3} &
  \textbf{GPT-5.2} &
  \textbf{Qwen3-235B} &
  \textbf{Mistral-Large} &
  \textbf{Claude-4.5-Son.} &
  \textbf{Qwen2.5-72B} &
  \textbf{Llama-3.3-70B} &
  \textbf{GPT-4o} \\
  \midrule
  \multirow{2}{*}{\textbf{Reading}} & Trace & 20 & 20 & 20 & 20 & 20 & 20 & 20 & 20 & 20 & 20 & 19 \\
   & Transpose & 20 & 20 & 20 & 20 & 20 & 20 & 20 & 20 & 18 & 19 & 20 \\
  \multicolumn{3}{l}{\textit{\textbf{Level Accuracy (\%)}}} & \textbf{100.0} & \textbf{100.0} & \textbf{100.0} & \textbf{100.0} & \textbf{100.0} & \textbf{100.0} & \textbf{100.0} & \textbf{95.0} & \textbf{97.5} & \textbf{97.5} \\
  \addlinespace[2pt]
  \multirow{3}{*}{\textbf{Arithmetic}} & Matrix-Vector & 20 & 20 & 20 & 20 & 20 & 19 & 20 & 20 & 19 & 20 & 20 \\
   & Multiplication & 20 & 20 & 20 & 20 & 20 & 19 & 20 & 20 & 19 & 19 & 20 \\
   & Matrix Power & 20 & 20 & 20 & 20 & 20 & 18 & 20 & 20 & 17 & 18 & 20 \\
  \multicolumn{3}{l}{\textit{\textbf{Level Accuracy (\%)}}} & \textbf{100.0} & \textbf{100.0} & \textbf{100.0} & \textbf{100.0} & \textbf{93.3} & \textbf{100.0} & \textbf{100.0} & \textbf{91.7} & \textbf{95.0} & \textbf{100.0} \\
  \addlinespace[2pt]
  \multirow{2}{*}{\textbf{Sequential}} & Nullity & 20 & 20 & 20 & 20 & 20 & 20 & 18 & 16 & 17 & 12 & 13 \\
   & Rank & 20 & 20 & 20 & 20 & 20 & 19 & 17 & 18 & 13 & 11 & 12 \\
  \multicolumn{3}{l}{\textit{\textbf{Level Accuracy (\%)}}} & \textbf{100.0} & \textbf{100.0} & \textbf{100.0} & \textbf{100.0} & \textbf{97.5} & \textbf{87.5} & \textbf{85.0} & \textbf{75.0} & \textbf{57.5} & \textbf{62.5} \\
  \addlinespace[2pt]
  \multirow{1}{*}{\textbf{Deep Recursive}} & Determinant & 50 & 50 & 50 & 47 & 50 & 49 & 47 & 39 & 13 & 43 & 11 \\
  \multicolumn{3}{l}{\textit{\textbf{Level Accuracy (\%)}}} & \textbf{100.0} & \textbf{100.0} & \textbf{94.0} & \textbf{100.0} & \textbf{98.0} & \textbf{94.0} & \textbf{78.0} & \textbf{26.0} & \textbf{86.0} & \textbf{22.0} \\
  \addlinespace[2pt]
  \multirow{1}{*}{\textbf{Built on Det.}} & Eigenvalues & 30 & 24 & 15 & 7 & 8 & 7 & 4 & 5 & 2 & 0 & 5 \\
  \multicolumn{3}{l}{\textit{\textbf{Level Accuracy (\%)}}} & \textbf{80.0} & \textbf{50.0} & \textbf{23.3} & \textbf{26.7} & \textbf{23.3} & \textbf{13.3} & \textbf{16.7} & \textbf{6.7} & \textbf{0.0} & \textbf{16.7} \\
  \addlinespace[2pt]
  \midrule
  \textbf{Total Correct} & & \textbf{220} & \textbf{214} & \textbf{205} & \textbf{194} & \textbf{198} & \textbf{191} & \textbf{186} & \textbf{178} & \textbf{138} & \textbf{162} & \textbf{140} \\
  \multicolumn{3}{l}{\textbf{Overall Model Accuracy (\%)}} & \textbf{97.3} & \textbf{93.2} & \textbf{88.2} & \textbf{90.0} & \textbf{86.8} & \textbf{84.5} & \textbf{80.9} & \textbf{62.7} & \textbf{73.6} & \textbf{63.6} \\
  \bottomrule
  \end{tabular}}
\end{table}

\begin{table}[p]
  \centering
  \caption{Complete accuracy table --- 5$\times$5 matrices (220 questions, 10 models). Tier~1: OpenAI-o1, Gemini-3.0-Pro, DeepSeek-V3, Qwen3-235B. Tier~2: GPT-5.2, Mistral-Large. Tier~3: Claude-4.5-Sonnet, Qwen2.5-72B, Llama-3.3-70B, GPT-4o.}
  \label{tab:c3}
  \renewcommand{\arraystretch}{1.15}
  \resizebox{\linewidth}{!}{%
  \begin{tabular}{ll r rrrrrrrrrr}
  \toprule
  \textbf{Level} & \textbf{Category} & \textbf{Max} &
  \textbf{OpenAI-o1} &
  \textbf{Gemini-3.0-Pro} &
  \textbf{DeepSeek-V3} &
  \textbf{GPT-5.2} &
  \textbf{Qwen3-235B} &
  \textbf{Mistral-Large} &
  \textbf{Claude-4.5-Son.} &
  \textbf{Qwen2.5-72B} &
  \textbf{Llama-3.3-70B} &
  \textbf{GPT-4o} \\
  \midrule
  \multirow{2}{*}{\textbf{Reading}} & Trace & 20 & 20 & 20 & 20 & 20 & 20 & 20 & 20 & 20 & 20 & 20 \\
   & Transpose & 20 & 20 & 20 & 20 & 20 & 20 & 20 & 20 & 19 & 20 & 20 \\
  \multicolumn{3}{l}{\textit{\textbf{Level Accuracy (\%)}}} & \textbf{100.0} & \textbf{100.0} & \textbf{100.0} & \textbf{100.0} & \textbf{100.0} & \textbf{100.0} & \textbf{100.0} & \textbf{97.5} & \textbf{100.0} & \textbf{100.0} \\
  \addlinespace[2pt]
  \multirow{3}{*}{\textbf{Arithmetic}} & Matrix-Vector & 20 & 20 & 20 & 20 & 20 & 20 & 20 & 20 & 16 & 20 & 20 \\
   & Multiplication & 20 & 20 & 20 & 20 & 20 & 15 & 20 & 20 & 12 & 19 & 19 \\
   & Matrix Power & 20 & 20 & 20 & 20 & 20 & 20 & 19 & 20 & 19 & 11 & 19 \\
  \multicolumn{3}{l}{\textit{\textbf{Level Accuracy (\%)}}} & \textbf{100.0} & \textbf{100.0} & \textbf{100.0} & \textbf{100.0} & \textbf{91.7} & \textbf{98.3} & \textbf{100.0} & \textbf{78.3} & \textbf{83.3} & \textbf{96.7} \\
  \addlinespace[2pt]
  \multirow{2}{*}{\textbf{Sequential}} & Nullity & 20 & 20 & 20 & 20 & 19 & 20 & 14 & 15 & 11 & 13 & 14 \\
   & Rank & 20 & 20 & 20 & 20 & 19 & 20 & 16 & 16 & 10 & 11 & 13 \\
  \multicolumn{3}{l}{\textit{\textbf{Level Accuracy (\%)}}} & \textbf{100.0} & \textbf{100.0} & \textbf{100.0} & \textbf{95.0} & \textbf{100.0} & \textbf{75.0} & \textbf{77.5} & \textbf{52.5} & \textbf{60.0} & \textbf{67.5} \\
  \addlinespace[2pt]
  \multirow{1}{*}{\textbf{Deep Recursive}} & Determinant & 50 & 50 & 50 & 37 & 27 & 45 & 18 & 6 & 0 & 2 & 0 \\
  \multicolumn{3}{l}{\textit{\textbf{Level Accuracy (\%)}}} & \textbf{100.0} & \textbf{100.0} & \textbf{74.0} & \textbf{54.0} & \textbf{90.0} & \textbf{36.0} & \textbf{12.0} & \textbf{0.0} & \textbf{4.0} & \textbf{0.0} \\
  \addlinespace[2pt]
  \multirow{1}{*}{\textbf{Built on Det.}} & Eigenvalues & 30 & 1 & 0 & 3 & 0 & 0 & 0 & 0 & 0 & 0 & 0 \\
  \multicolumn{3}{l}{\textit{\textbf{Level Accuracy (\%)}}} & \textbf{3.3} & \textbf{0.0} & \textbf{10.0} & \textbf{0.0} & \textbf{0.0} & \textbf{0.0} & \textbf{0.0} & \textbf{0.0} & \textbf{0.0} & \textbf{0.0} \\
  \addlinespace[2pt]
  \midrule
  \textbf{Total Correct} & & \textbf{220} & \textbf{191} & \textbf{190} & \textbf{180} & \textbf{165} & \textbf{180} & \textbf{147} & \textbf{137} & \textbf{107} & \textbf{116} & \textbf{125} \\
  \multicolumn{3}{l}{\textbf{Overall Model Accuracy (\%)}} & \textbf{87.0} & \textbf{86.0} & \textbf{82.0} & \textbf{75.0} & \textbf{82.0} & \textbf{67.0} & \textbf{62.0} & \textbf{49.0} & \textbf{53.0} & \textbf{57.0} \\
  \bottomrule
  \end{tabular}}
\end{table}


\section{Format Sensitivity Analysis}
\label{app:format}

Format sensitivity was evaluated across three input representations ---
LaTeX (standard bmatrix notation), List (nested Python-style list), and
Tabular (pipe-delimited grid) --- at $4{\times}4$ and $5{\times}5$ matrix
dimensions only; the $3{\times}3$ dimension was excluded as pilot evaluation
confirmed near-universal ceiling performance across all formats at that scale,
making format the least interesting variable to study. $n{=}200$ questions per
model per format across 9 subcategories.

At $4{\times}4$ format sensitivity is negligible across all tiers.
At $5{\times}5$ it is strictly complexity-gated:
\begin{itemize}
  \item \textbf{Tier~1 models} remain format-robust. Gemini achieves 30/30
        under all formats at 5$\times$5.
  \item \textbf{Tier~2 models} show divergent model-specific sensitivity.
        GPT-5.2 incurs a \textbf{LaTeX penalty} (14/30 LaTeX vs 26--27/30
        Tabular/List); Mistral shows the opposite. These are opposing effects,
        not systematic format bias.
  \item \textbf{Tier~3 models} collapse completely regardless of format ---
        GPT-4o achieves 0/30 and Llama at most 1/30 across all three formats.
        The dimensional boundary, not input notation, is the binding constraint.
\end{itemize}

The key implication: format sensitivity is a \textbf{second-order effect} that
manifests only when working memory is partially saturated (Tier~2 at 5$\times$5).
When sufficient (Tier~1) or fully saturated (Tier~3), format is irrelevant.

\begin{figure}[t]
  \centering
  \includegraphics[width=\linewidth]{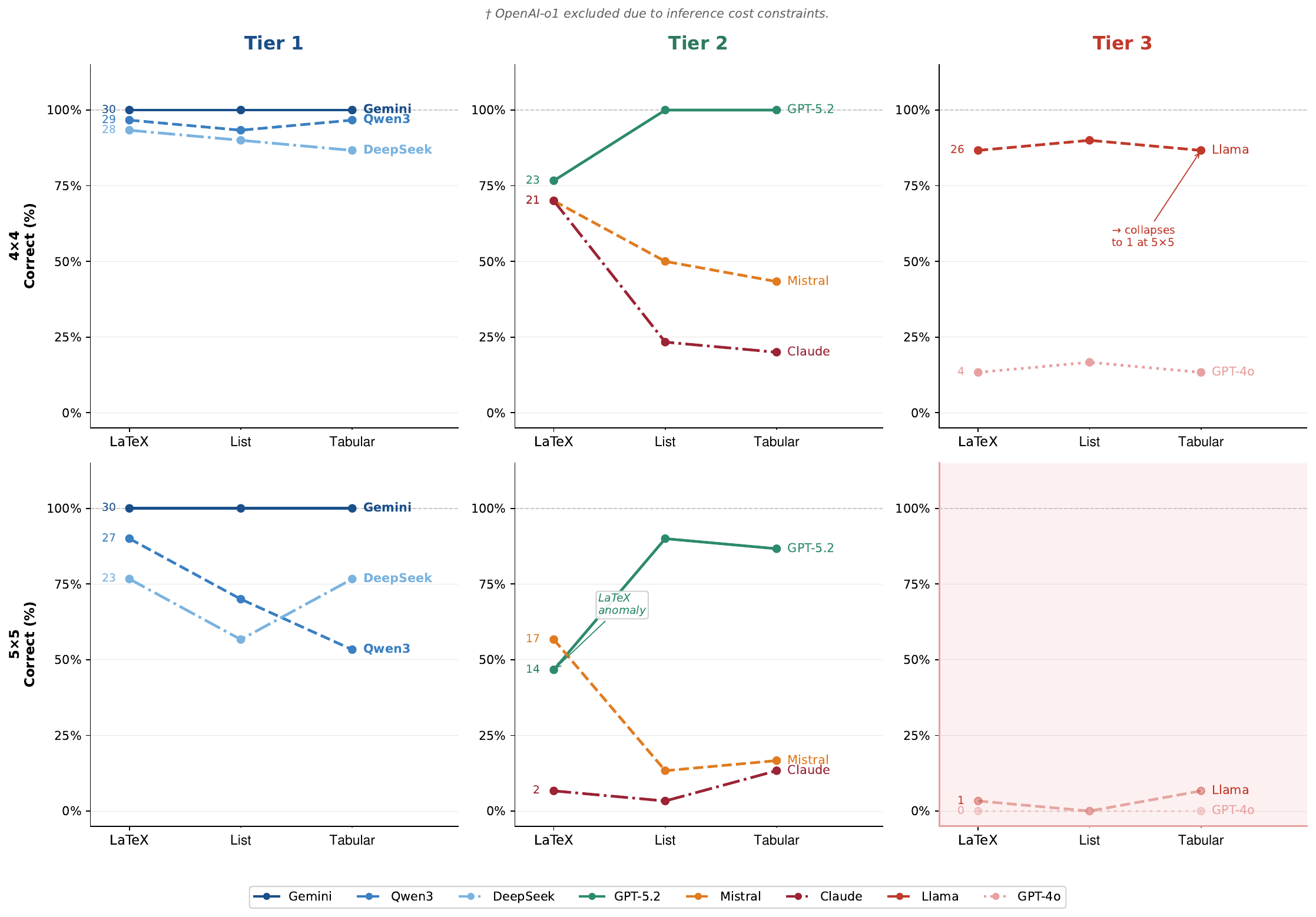}
  \caption{\textit{Format sensitivity across matrix dimensions and model tiers
    (determinant subcategory, $n{=}30$ per model per format). At $4{\times}4$
    (top row) format variance is minimal. At $5{\times}5$ (bottom row) format
    sensitivity emerges but is strictly complexity-gated. Red-shaded panel
    indicates complete Tier~3 collapse. $\dagger$~OpenAI-o1 excluded due to
    inference cost constraints.}}
  \label{fig:d1}
\end{figure}

\begin{table}[p]
  \centering
  \caption{Format sensitivity raw correct counts --- 4$\times$4 matrices (200 questions per model per format, 9 subcategories). Three sub-columns per model: LaTeX, List, Tabular. All counts from a fully enumerated evaluation. $\dagger$ OpenAI-o1 excluded.}
  \label{tab:d1a}
  \renewcommand{\arraystretch}{1.1}
  \resizebox{\linewidth}{!}{%
  \begin{tabular}{l r ccc ccc ccc ccc ccc ccc}
  \toprule
  \textbf{Category} & \textbf{Max} & \multicolumn{3}{c}{\textbf{Gemini}} & \multicolumn{3}{c}{\textbf{Qwen 235B}} & \multicolumn{3}{c}{\textbf{DeepSeek}} & \multicolumn{3}{c}{\textbf{GPT-5.2}} & \multicolumn{3}{c}{\textbf{Mistral}} & \multicolumn{3}{c}{\textbf{Claude 4.5}} \\
  \cmidrule(lr){3--5} \cmidrule(lr){6--8} \cmidrule(lr){9--11} \cmidrule(lr){12--14} \cmidrule(lr){15--17} \cmidrule(lr){18--20}
  & & \textbf{LaTeX} & \textbf{List} & \textbf{Tab.}\textbf{LaTeX} & \textbf{List} & \textbf{Tab.}\textbf{LaTeX} & \textbf{List} & \textbf{Tab.}\textbf{LaTeX} & \textbf{List} & \textbf{Tab.}\textbf{LaTeX} & \textbf{List} & \textbf{Tab.}\textbf{LaTeX} & \textbf{List} & \textbf{Tab.} \\
  \midrule
  \textbf{Determinant} & \textbf{30} & 30 & 30 & 30 & 28 & 30 & 30 & 28 & 27 & 26 & 23 & 30 & 30 & 26 & 28 & 26 & 19 & 21 & 7 \\
  \textbf{Eigenvalues} & \textbf{30} & 15 & 10 & 10 & 5 & 8 & 5 & 6 & 9 & 10 & 8 & 16 & 13 & 5 & 5 & 6 & 5 & 5 & 5 \\
  \textbf{Multiplication} & \textbf{20} & 20 & 20 & 20 & 19 & 20 & 18 & 20 & 20 & 20 & 20 & 20 & 20 & 17 & 20 & 20 & 20 & 20 & 20 \\
  \textbf{Rank} & \textbf{20} & 20 & 20 & 20 & 7 & 20 & 20 & 20 & 20 & 20 & 20 & 20 & 20 & 6 & 18 & 16 & 17 & 17 & 20 \\
  \textbf{Trace} & \textbf{20} & 20 & 20 & 20 & 20 & 20 & 20 & 20 & 20 & 20 & 20 & 20 & 20 & 20 & 20 & 20 & 20 & 20 & 20 \\
  \textbf{Matrix-Vector} & \textbf{20} & 20 & 20 & 20 & 19 & 19 & 20 & 20 & 20 & 20 & 20 & 20 & 20 & 18 & 20 & 20 & 20 & 20 & 20 \\
  \textbf{Nullity} & \textbf{20} & 20 & 20 & 20 & 15 & 20 & 20 & 20 & 19 & 20 & 20 & 20 & 20 & 8 & 17 & 14 & 19 & 15 & 15 \\
  \textbf{Transpose} & \textbf{20} & 20 & 20 & 20 & 20 & 20 & 20 & 20 & 20 & 20 & 20 & 20 & 19 & 20 & 20 & 20 & 20 & 20 & 20 \\
  \textbf{Matrix Power} & \textbf{20} & 20 & 15 & 15 & 20 & 19 & 19 & 20 & 20 & 20 & 20 & 13 & 14 & 19 & 20 & 19 & 20 & 20 & 20 \\
  \midrule
  \textbf{TOTAL} & \textbf{200} & \textbf{185} & \textbf{175} & \textbf{172} & \textbf{170} & \textbf{175} & \textbf{172} & \textbf{171} & \textbf{179} & \textbf{176} & \textbf{172} & \textbf{175} & \textbf{174} & \textbf{165} & \textbf{163} & \textbf{167} & \textbf{163} & \textbf{159} & \textbf{159} \\
  \bottomrule
  \end{tabular}}
\end{table}

\begin{table}[p]
  \centering
  \caption{Format sensitivity raw correct counts --- 4$\times$4 matrices, Tier 3 models.}
  \label{tab:d1b}
  \renewcommand{\arraystretch}{1.1}
  \resizebox{\linewidth}{!}{%
  \begin{tabular}{l r ccc ccc ccc}
  \toprule
  \textbf{Category} & \textbf{Max} & \multicolumn{3}{c}{\textbf{GPT-4o}} & \multicolumn{3}{c}{\textbf{Llama}} & \multicolumn{3}{c}{\textbf{Qwen2.5 72B}} \\
  \cmidrule(lr){3--5} \cmidrule(lr){6--8} \cmidrule(lr){9--11}
  & & \textbf{LaTeX} & \textbf{List} & \textbf{Tab.}\textbf{LaTeX} & \textbf{List} & \textbf{Tab.}\textbf{LaTeX} & \textbf{List} & \textbf{Tab.} \\
  \midrule
  \textbf{Determinant} & \textbf{30} & 5 & 5 & 4 & 26 & 27 & 26 & 6 & 7 & 28 \\
  \textbf{Eigenvalues} & \textbf{30} & 5 & 5 & 5 & 4 & 4 & 4 & 5 & 5 & 5 \\
  \textbf{Multiplication} & \textbf{20} & 20 & 18 & 20 & 20 & 19 & 17 & 20 & 19 & 19 \\
  \textbf{Rank} & \textbf{20} & 11 & 11 & 6 & 19 & 6 & 11 & 17 & 9 & 10 \\
  \textbf{Trace} & \textbf{20} & 20 & 20 & 19 & 20 & 20 & 20 & 20 & 20 & 20 \\
  \textbf{Matrix-Vector} & \textbf{20} & 20 & 20 & 19 & 20 & 19 & 20 & 19 & 20 & 19 \\
  \textbf{Nullity} & \textbf{20} & 15 & 12 & 9 & 20 & 12 & 11 & 16 & 17 & 17 \\
  \textbf{Transpose} & \textbf{20} & 20 & 20 & 20 & 20 & 20 & 19 & 19 & 19 & 20 \\
  \textbf{Matrix Power} & \textbf{20} & 18 & 20 & 17 & 17 & 17 & 16 & 20 & 20 & 18 \\
  \midrule
  \textbf{TOTAL} & \textbf{200} & \textbf{134} & \textbf{131} & \textbf{119} & \textbf{143} & \textbf{146} & \textbf{135} & \textbf{135} & \textbf{134} & \textbf{132} \\
  \bottomrule
  \end{tabular}}
\end{table}

\begin{table}[p]
  \centering
  \caption{Format sensitivity raw correct counts --- 5$\times$5 matrices (200 questions per model per format, 9 subcategories). Eigenvalue row shows near-universal zero across all models and formats, confirming eigenvalue collapse is format-invariant.}
  \label{tab:d2a}
  \renewcommand{\arraystretch}{1.1}
  \resizebox{\linewidth}{!}{%
  \begin{tabular}{l r ccc ccc ccc ccc ccc ccc}
  \toprule
  \textbf{Category} & \textbf{Max} & \multicolumn{3}{c}{\textbf{Gemini}} & \multicolumn{3}{c}{\textbf{Qwen 235B}} & \multicolumn{3}{c}{\textbf{DeepSeek}} & \multicolumn{3}{c}{\textbf{GPT-5.2}} & \multicolumn{3}{c}{\textbf{Mistral}} & \multicolumn{3}{c}{\textbf{Claude 4.5}} \\
  \cmidrule(lr){3--5} \cmidrule(lr){6--8} \cmidrule(lr){9--11} \cmidrule(lr){12--14} \cmidrule(lr){15--17} \cmidrule(lr){18--20}
  & & \textbf{LaTeX} & \textbf{List} & \textbf{Tab.}\textbf{LaTeX} & \textbf{List} & \textbf{Tab.}\textbf{LaTeX} & \textbf{List} & \textbf{Tab.}\textbf{LaTeX} & \textbf{List} & \textbf{Tab.}\textbf{LaTeX} & \textbf{List} & \textbf{Tab.}\textbf{LaTeX} & \textbf{List} & \textbf{Tab.} \\
  \midrule
  \textbf{Determinant} & \textbf{30} & 30 & 30 & 30 & 27 & 23 & 21 & 23 & 17 & 23 & 14 & 27 & 26 & 17 & 4 & 5 & 2 & 4 & 2 \\
  \textbf{Eigenvalues} & \textbf{30} & 0 & 0 & 0 & 0 & 0 & 0 & 3 & 0 & 0 & 0 & 0 & 0 & 0 & 0 & 0 & 0 & 0 & 0 \\
  \textbf{Multiplication} & \textbf{20} & 20 & 20 & 20 & 16 & 16 & 16 & 20 & 20 & 19 & 19 & 19 & 18 & 20 & 15 & 19 & 19 & 20 & 19 \\
  \textbf{Rank} & \textbf{20} & 20 & 20 & 20 & 20 & 20 & 20 & 20 & 20 & 20 & 19 & 19 & 20 & 15 & 13 & 15 & 16 & 12 & 10 \\
  \textbf{Trace} & \textbf{20} & 20 & 20 & 20 & 20 & 20 & 20 & 20 & 20 & 20 & 20 & 20 & 20 & 20 & 20 & 18 & 20 & 20 & 20 \\
  \textbf{Matrix-Vector} & \textbf{20} & 20 & 20 & 20 & 20 & 17 & 19 & 20 & 20 & 20 & 20 & 20 & 20 & 20 & 20 & 20 & 20 & 20 & 20 \\
  \textbf{Nullity} & \textbf{20} & 20 & 20 & 20 & 20 & 20 & 18 & 20 & 20 & 20 & 19 & 15 & 17 & 14 & 9 & 11 & 15 & 12 & 13 \\
  \textbf{Transpose} & \textbf{20} & 20 & 20 & 20 & 20 & 18 & 19 & 20 & 20 & 20 & 20 & 20 & 20 & 20 & 19 & 18 & 20 & 20 & 20 \\
  \textbf{Matrix Power} & \textbf{20} & 20 & 20 & 20 & 20 & 14 & 17 & 20 & 16 & 17 & 20 & 18 & 20 & 19 & 17 & 17 & 20 & 20 & 20 \\
  \midrule
  \textbf{TOTAL} & \textbf{200} & \textbf{170} & \textbf{170} & \textbf{170} & \textbf{163} & \textbf{148} & \textbf{150} & \textbf{166} & \textbf{151} & \textbf{159} & \textbf{151} & \textbf{158} & \textbf{161} & \textbf{145} & \textbf{117} & \textbf{123} & \textbf{132} & \textbf{121} & \textbf{117} \\
  \bottomrule
  \end{tabular}}
  \\[4pt]\footnotesize\textit{$\dagger$ OpenAI-o1 excluded. Format sensitivity covers 9 models $\times$ 3 formats $\times$ 200 questions = 5{,}400 evaluations per dimension.}
\end{table}

\begin{table}[p]
  \centering
  \caption{Format sensitivity raw correct counts --- 5$\times$5 matrices, Tier 3 models.}
  \label{tab:d2b}
  \renewcommand{\arraystretch}{1.1}
  \resizebox{\linewidth}{!}{%
  \begin{tabular}{l r ccc ccc ccc}
  \toprule
  \textbf{Category} & \textbf{Max} & \multicolumn{3}{c}{\textbf{GPT-4o}} & \multicolumn{3}{c}{\textbf{Llama}} & \multicolumn{3}{c}{\textbf{Qwen2.5 72B}} \\
  \cmidrule(lr){3--5} \cmidrule(lr){6--8} \cmidrule(lr){9--11}
  & & \textbf{LaTeX} & \textbf{List} & \textbf{Tab.}\textbf{LaTeX} & \textbf{List} & \textbf{Tab.}\textbf{LaTeX} & \textbf{List} & \textbf{Tab.} \\
  \midrule
  \textbf{Determinant} & \textbf{30} & 0 & 0 & 0 & 1 & 0 & 2 & 1 & 0 & 1 \\
  \textbf{Eigenvalues} & \textbf{30} & 0 & 0 & 0 & 0 & 0 & 0 & 0 & 0 & 0 \\
  \textbf{Multiplication} & \textbf{20} & 19 & 16 & 18 & 19 & 13 & 18 & 9 & 10 & 10 \\
  \textbf{Rank} & \textbf{20} & 13 & 10 & 8 & 11 & 9 & 12 & 10 & 9 & 7 \\
  \textbf{Trace} & \textbf{20} & 20 & 19 & 19 & 20 & 20 & 20 & 20 & 20 & 20 \\
  \textbf{Matrix-Vector} & \textbf{20} & 20 & 13 & 12 & 20 & 13 & 13 & 16 & 10 & 10 \\
  \textbf{Nullity} & \textbf{20} & 14 & 7 & 13 & 13 & 8 & 9 & 11 & 12 & 10 \\
  \textbf{Transpose} & \textbf{20} & 20 & 20 & 20 & 20 & 17 & 18 & 19 & 19 & 20 \\
  \textbf{Matrix Power} & \textbf{20} & 19 & 15 & 18 & 11 & 6 & 5 & 19 & 16 & 16 \\
  \midrule
  \textbf{TOTAL} & \textbf{200} & \textbf{125} & \textbf{100} & \textbf{108} & \textbf{115} & \textbf{86} & \textbf{97} & \textbf{105} & \textbf{96} & \textbf{94} \\
  \bottomrule
  \end{tabular}}
  \\[4pt]\footnotesize\textit{$\dagger$ OpenAI-o1 excluded. Format sensitivity covers 9 models $\times$ 3 formats $\times$ 200 questions = 5{,}400 evaluations per dimension.}
\end{table}


\maketitle

\section{Judge Pipeline Validation}
\label{app:judge-pipeline}
\label{app:judge}

The forensic classification pipeline operates in two independent stages.
Both stages use \textbf{Gemini 3.1 Pro Preview} at \textbf{temperature 0.0}
for deterministic output.

\subsection{Build Judge (Stage 1 --- First-Pass Classification)}
\label{app:build-judge}

The Build Judge acts as a forensic auditor. For each failed response:

\begin{itemize}

  \item For all subcategories except eigenvalues, \textbf{the Build Judge}
    independently computes the correct solution from scratch before reading
    the model response. Gemini-3.0-Pro achieves perfect accuracy on all
    non-eigenvalue subcategories across all three matrix dimensions,
    confirming it as a reliable computation oracle for the judge role. All
    judge outputs are additionally cross-validated against the SymPy-certified
    ground truth \citep{meurer2017sympy} as a consistency check. For eigenvalues,
    the Build Judge receives the SymPy-certified answer directly as its
    reference and focuses solely on tracing where and how the model's
    reasoning diverged from the correct solution path.

  \item Traces the model response \textbf{step-by-step} to identify the
    \textbf{first erroneous value}, not just the final wrong answer.

  \item Applies the \textbf{Magnitude Rule} as the primary boundary enforcer
    between \textsc{sign\_error} and \textsc{arithmetic}: if
    $\lvert\text{wrong}\rvert = \lvert\text{correct}\rvert$ (element-wise
    for matrices/vectors) $\rightarrow$ \textsc{sign\_error}; if magnitudes
    differ $\rightarrow$ \textsc{arithmetic}.

  \item Records \textbf{Solution\_Strategy} (e.g., cofactor expansion
    vs.\ Gaussian elimination) alongside the error tag.

  \item Outputs \textbf{First\_Error\_Step} and
    \textbf{First\_Error\_Description} for full forensic traceability.

  \item Requires a \textbf{confidence level} (\textsc{high} /
    \textsc{medium} / \textsc{low}) for every classification.

\end{itemize}

\subsection{Validate Judge (Stage 2 --- Independent Verification)}
\label{app:validate-judge}

Stage 2 is not an independent classifier but an \textbf{independent
verification step} designed to stress-test the Build Judge's proposed
hypothesis against strict boundary rules, minimising false positives:

\begin{itemize}

  \item A \textbf{second Gemini call} is made with no access to the Build
    Judge's reasoning or forensic observation --- only the error tag and
    original model response are passed.

  \item A \textbf{tag-specific verification question} is injected per
    classification (see Table~E.1), enforcing a targeted diagnostic check.

  \item The Validate Judge can \textbf{confirm} (\textsc{true}),
    \textbf{correct} (\textsc{false} + corrected tag), or \textbf{flag}
    (\textsc{needs\_review}).

  \item Final tag cascade: if validated $=$ \textsc{false} and
    \texttt{corrected\_tag} non-empty $\rightarrow$ \texttt{corrected\_tag}
    becomes \textsc{final\_tag}; otherwise original \texttt{Error\_Tag} is
    retained.

\end{itemize}

Overall Validate Judge agreement was 83.0\% at $3{\times}3$, 83.2\% at
$4{\times}4$, and 82.0\% at $5{\times}5$. Per-tag agreement rates are
reported in Table~E.3.

\subsection{Human Batch Review (Stage 3 --- Taxonomy Refinement)}
\label{app:human-review}

In a third stage, systematic disagreements between Build and Validate Judge
outputs were reviewed manually at batch level using a structured spreadsheet
audit. Recurring disagreement patterns identified ambiguities in the taxonomy
definitions, which were resolved by refining the judge prompts iteratively.
This process produced the final taxonomy --- for example, deprecating the
overly broad \texttt{Fluency\_Masking} tag in favour of
\texttt{Silent\_Sign\_Flip}, and splitting \textsc{transcription} into
\textsc{input\_transcription} and \textsc{carry\_down\_error}.

\begin{table}[H]
\centering
\small
\caption{\textit{Tag-specific verification questions injected into the
Validate Judge prompt. For each primary tag assigned by the Build Judge, the
Validate Judge receives the corresponding verification question and boundary
rule before confirming or correcting the classification.}}
\label{tab:E1}
\resizebox{\textwidth}{!}{%
\begin{tabular}{@{} p{4.2cm} p{5.6cm} p{3.8cm} @{}}
\toprule
\textbf{Primary Tag} &
\textbf{Validate Judge Verification Question} &
\textbf{Boundary Rule} \\
\midrule
\textbf{SIGN\_ERROR} &
Does the magnitude of the wrong answer equal the magnitude of the correct
answer? For scalar answers: $\lvert\text{wrong}\rvert =
\lvert\text{correct}\rvert$. For vectors and matrices: every corresponding
element must match in absolute magnitude ---
$\lvert\text{wrong}_{ij}\rvert = \lvert\text{correct}_{ij}\rvert$ for all
$i,j$. If yes $\rightarrow$ \textsc{sign\_error}. If magnitudes differ
$\rightarrow$ reclassify as \textsc{arithmetic}. &
\textit{Magnitude Rule (primary boundary enforcer between
\textsc{sign\_error} and \textsc{arithmetic}).} \\
\midrule
\textbf{ARITHMETIC} &
Does the wrong answer differ in absolute magnitude from the correct answer,
ruling out a pure sign flip? For matrices/vectors: at least one element
differs in magnitude. &
\textit{Magnitude Rule inverse. Must confirm
$\lvert\text{wrong}\rvert \neq \lvert\text{correct}\rvert$.} \\
\midrule
\textbf{HALLUCINATION} &
Does the model explicitly abandon computation, fabricate a final answer
without valid working, or produce fewer than 2 meaningful computation steps
before asserting a result? &
\textit{Truncation Precheck: \textsc{generation\_truncation} must be ruled
out first --- check if response ends abruptly vs.\ contains an explicit
abandonment statement.} \\
\midrule
\textbf{METHOD\_FAIL} &
Does the model apply a fundamentally incorrect algorithm from the very first
step --- not merely make an arithmetic or sign error mid-computation? &
\textit{Must confirm wrong method from step 1. Execution error within a
correct method is not \textsc{method\_fail}.} \\
\midrule
\textbf{INPUT\_TRANSCRIPTION} &
Does the model misread or incorrectly copy a matrix entry before any
arithmetic begins --- i.e., is the error at input parsing, not during
computation? &
\textit{Error must occur before first arithmetic operation.} \\
\midrule
\textbf{CARRY\_DOWN\_ERROR} &
Is the value correctly stated at step $N$ but then incorrectly copied into
step $N{+}1$, with no arithmetic error at step $N$ itself? &
\textit{Must distinguish from \textsc{memory\_loss}: error must be at the
immediately following step ($N{+}1$), not a distant later step.} \\
\midrule
\textbf{MEMORY\_LOSS} &
Is the value correctly stated at step $N$ but recalled incorrectly at a much
later step $M$ ($M \gg N$), with correct copying at all intermediate steps? &
\textit{Must confirm temporal distance. \textsc{carry\_down\_error} if error
is at $N{+}1$ only.} \\
\midrule
\textbf{GENERATION\_TRUNCATION} &
Does the response end abruptly without a final answer, consistent with a
token limit or generation cutoff rather than an explicit reasoning failure or
abandonment? &
\textit{\textsc{hallucination} boundary: truncation has no abandonment
language; \textsc{hallucination} has explicit meta-statements.} \\
\midrule
\textbf{FORMATTING\_MISMATCH} &
Are all computed numerical values correct but presented in a format that
fails the output structure requirement? &
\textit{All values must be numerically correct. Any numerical error overrides
this tag.} \\
\midrule
\textbf{OTHER\_UNMAPPED} &
Does the failure genuinely not fit any of the nine tags above after careful
step-by-step consideration? &
\textit{Last resort only. Requires explicit elimination of all nine tags.} \\
\bottomrule
\end{tabular}%
}
\end{table}

\begin{table}[H]
\centering
\small
\caption{\textit{Error tag and sub-tag applicability by problem subcategory.
A checkmark ($\checkmark$) indicates the tag can structurally occur in that
subcategory; a dash (---) indicates structural impossibility confirmed by the
pipeline taxonomy. Ten primary tags (rows 1--10) apply across all
subcategories with the exception of \textsc{trace}, which has no
\textsc{sign\_error}, \textsc{hallucination}, or
\textsc{carry\_down\_error} due to its purely additive nature. Four
eigenvalue-specific tags ($\dagger$) apply exclusively to the \textsc{eig}
subcategory, reflecting unique failure modes in characteristic polynomial
expansion.}}
\label{tab:E2}
\resizebox{\linewidth}{!}{%
\begin{tabular}{@{} r l l *{9}{c} @{}}
\toprule
\textbf{\#} & \textbf{Tag} & \textbf{Sub-tag} &
\textbf{DET} & \textbf{EIG} & \textbf{RANK} & \textbf{NULL} &
\textbf{MULT} & \textbf{POW} & \textbf{VEC} & \textbf{TRANS} &
\textbf{TRACE} \\
\midrule
\multicolumn{12}{@{}l}{\textbf{1 \quad SIGN\_ERROR}} \\
& sub tag & \textit{Alternating\_Drift}
  & $\checkmark$ & $\checkmark$ & ---          & ---          & ---          & ---          & ---          & ---          & --- \\
& & \textit{Cofactor\_Neglect}
  & $\checkmark$ & $\checkmark$ & ---          & ---          & ---          & ---          & ---          & ---          & --- \\
& & \textit{Parity\_Sign\_Error}
  & $\checkmark$ & $\checkmark$ & ---          & ---          & ---          & ---          & ---          & ---          & --- \\
& & \textit{Product\_Sign\_Error}
  & $\checkmark$ & $\checkmark$ & $\checkmark$ & $\checkmark$ & $\checkmark$ & $\checkmark$ & $\checkmark$ & ---          & --- \\
& & \textit{Double\_Negative\_Trap}
  & $\checkmark$ & $\checkmark$ & $\checkmark$ & $\checkmark$ & $\checkmark$ & $\checkmark$ & $\checkmark$ & ---          & --- \\
& & \textit{Rule\_Interference}
  & $\checkmark$ & $\checkmark$ & $\checkmark$ & $\checkmark$ & $\checkmark$ & $\checkmark$ & $\checkmark$ & ---          & --- \\
& & \textit{Operation\_Direction}
  & $\checkmark$ & $\checkmark$ & $\checkmark$ & $\checkmark$ & $\checkmark$ & $\checkmark$ & $\checkmark$ & ---          & --- \\
& & \textit{Silent\_Sign\_Flip}
  & $\checkmark$ & $\checkmark$ & $\checkmark$ & $\checkmark$ & $\checkmark$ & $\checkmark$ & $\checkmark$ & ---          & --- \\
\multicolumn{12}{@{}l}{\textbf{2 \quad HALLUCINATION}} \\
& sub tag & \textit{Complete\_Collapse}
  & $\checkmark$ & $\checkmark$ & $\checkmark$ & $\checkmark$ & $\checkmark$ & $\checkmark$ & $\checkmark$ & $\checkmark$ & --- \\
& & \textit{Teleological\_Zeroing}
  & $\checkmark$ & $\checkmark$ & $\checkmark$ & $\checkmark$ & $\checkmark$ & $\checkmark$ & $\checkmark$ & $\checkmark$ & --- \\
& & \textit{Premature\_Assertion}
  & $\checkmark$ & $\checkmark$ & $\checkmark$ & $\checkmark$ & $\checkmark$ & $\checkmark$ & $\checkmark$ & $\checkmark$ & --- \\
& & \textit{Silent\_Omission}
  & $\checkmark$ & $\checkmark$ & $\checkmark$ & $\checkmark$ & $\checkmark$ & $\checkmark$ & $\checkmark$ & $\checkmark$ & --- \\
& & \textit{Ungrounded\_Guess}
  & $\checkmark$ & $\checkmark$ & $\checkmark$ & $\checkmark$ & $\checkmark$ & $\checkmark$ & $\checkmark$ & $\checkmark$ & --- \\
& & \textit{Spontaneous\_Insertion}
  & $\checkmark$ & $\checkmark$ & $\checkmark$ & $\checkmark$ & $\checkmark$ & $\checkmark$ & $\checkmark$ & $\checkmark$ & --- \\
& & \textit{Dimension\_Assertion}
  & ---          & ---          & ---          & ---          & ---          & ---          & $\checkmark$ & ---          & --- \\
\multicolumn{12}{@{}l}{\textbf{3 \quad METHOD\_FAIL}} \\
& sub tag & \textit{Operand\_Confusion}
  & ---          & ---          & ---          & ---          & $\checkmark$ & $\checkmark$ & $\checkmark$ & ---          & --- \\
& & \textit{Composition\_Rule\_Violation}
  & ---          & ---          & ---          & ---          & ---          & $\checkmark$ & ---          & $\checkmark$ & --- \\
& & \textit{(base --- no subtype)}
  & $\checkmark$ & $\checkmark$ & $\checkmark$ & $\checkmark$ & ---          & ---          & ---          & ---          & $\checkmark$ \\
\multicolumn{12}{@{}l}{\textbf{4 \quad ARITHMETIC}} \\
& & \textit{(no sub-tag)}
  & $\checkmark$ & $\checkmark$ & $\checkmark$ & $\checkmark$ & $\checkmark$ & $\checkmark$ & $\checkmark$ & $\checkmark$ & $\checkmark$ \\
\multicolumn{12}{@{}l}{\textbf{5 \quad INPUT\_TRANSCRIPTION}} \\
& & \textit{(no sub-tag)}
  & $\checkmark$ & $\checkmark$ & $\checkmark$ & $\checkmark$ & $\checkmark$ & $\checkmark$ & $\checkmark$ & $\checkmark$ & $\checkmark$ \\
\multicolumn{12}{@{}l}{\textbf{6 \quad CARRY\_DOWN\_ERROR}} \\
& & \textit{(no sub-tag)}
  & $\checkmark$ & $\checkmark$ & $\checkmark$ & $\checkmark$ & $\checkmark$ & $\checkmark$ & $\checkmark$ & ---          & --- \\
\multicolumn{12}{@{}l}{\textbf{7 \quad MEMORY\_LOSS}} \\
& & \textit{(no sub-tag)}
  & $\checkmark$ & $\checkmark$ & $\checkmark$ & $\checkmark$ & $\checkmark$ & $\checkmark$ & $\checkmark$ & $\checkmark$ & $\checkmark$ \\
\multicolumn{12}{@{}l}{\textbf{8 \quad GENERATION\_TRUNCATION}} \\
& & \textit{(no sub-tag)}
  & $\checkmark$ & $\checkmark$ & $\checkmark$ & $\checkmark$ & $\checkmark$ & $\checkmark$ & $\checkmark$ & $\checkmark$ & $\checkmark$ \\
\multicolumn{12}{@{}l}{\textbf{9 \quad FORMATTING\_MISMATCH}} \\
& & \textit{(no sub-tag)}
  & $\checkmark$ & $\checkmark$ & $\checkmark$ & $\checkmark$ & $\checkmark$ & $\checkmark$ & $\checkmark$ & $\checkmark$ & $\checkmark$ \\
\multicolumn{12}{@{}l}{\textbf{10 \quad OTHER\_UNMAPPED}} \\
& & \textit{(no sub-tag)}
  & $\checkmark$ & $\checkmark$ & $\checkmark$ & $\checkmark$ & $\checkmark$ & $\checkmark$ & $\checkmark$ & $\checkmark$ & $\checkmark$ \\
\midrule
\multicolumn{12}{@{}l}{\textbf{Eigenvalue-specific tags --- apply to EIG subcategory only $\dagger$}} \\
11 & GENERATION\_LOOP       & \textit{(no sub-tag)} & --- & $\checkmark$ & --- & --- & --- & --- & --- & --- & --- \\
12 & ALGEBRAIC\_PRECEDENCE  & \textit{(no sub-tag)} & --- & $\checkmark$ & --- & --- & --- & --- & --- & --- & --- \\
13 & FALSE\_VERIFICATION    & \textit{(no sub-tag)} & --- & $\checkmark$ & --- & --- & --- & --- & --- & --- & --- \\
14 & VARIABLE\_ENTANGLEMENT & \textit{(no sub-tag)} & --- & $\checkmark$ & --- & --- & --- & --- & --- & --- & --- \\
\bottomrule
\end{tabular}%
}

\smallskip
\noindent\textit{$\dagger$ Eigenvalue-specific tags
(\textsc{generation\_loop}, \textsc{algebraic\_precedence},
\textsc{false\_verification}, \textsc{variable\_entanglement}) are additional
primary tags that arise exclusively during the root-finding stage of
characteristic polynomial expansion. The benchmark abstract references ten
primary tags, which reflects the \textsc{base\_primary\_tags} applied across
all subcategories; these four are additional and eigenvalue-only.}
\end{table}

\subsection{Agreement Scope and Distributional Validity}
\label{app:agreement-scope}

\textit{Tag-level agreement rates in Table~E.3 vary substantially across
dimensions. This variation is expected and does not undermine the pipeline's
distributional findings for three reasons.}

\textit{First, the five tags that drive the distributional analysis in
Section~5 --- \textsc{sign\_error}, \textsc{hallucination},
\textsc{arithmetic}, \textsc{input\_transcription}, and
\textsc{method\_fail} --- collectively account for
\textbf{91.5\% of all classified failures}. These five tags achieve the
highest and most consistent agreement rates across all dimensions
(Table~E.3), confirming that the pipeline is most reliable precisely where
it matters most.}

\textit{Second, tags with ${<}70\%$ agreement at any dimension
(\textsc{generation\_truncation}, \textsc{memory\_loss},
\textsc{variable\_entanglement}, \textsc{algebraic\_precedence},
\textsc{other\_unmapped}, \textsc{generation\_loop}) are low-frequency tags
that collectively account for fewer than \textbf{8.5\% of failures}. Their
exclusion from distributional analysis in Section~5 is conservative by
design --- these tags are retained in the full taxonomy for completeness but
do not drive any quantitative claims in the paper.}

\textit{Third, the $3{\times}3$ dimension achieves 100\% human agreement,
and $4{\times}4$ achieves 96.7\%, confirming that pipeline reliability
degrades gracefully with complexity rather than collapsing. The $5{\times}5$
figure of 89.7\% reflects the genuine difficulty of classifying failures at
maximum recursive depth, not a systematic pipeline bias.}

\begin{table}[H]
\centering
\small
\caption{\textbf{\textit{Per-tag Validate Judge Agreement Rates}}}
\label{tab:E3}
\resizebox{\textwidth}{!}{%
\begin{tabular}{@{} l c c c p{5.0cm} @{}}
\toprule
\textit{\textbf{Tag}} &
\textit{\textbf{3$\times$3 Agree\%}} &
\textit{\textbf{4$\times$4 Agree\%}} &
\textit{\textbf{5$\times$5 Agree\%}} &
\textit{\textbf{Note}} \\
\midrule
\textit{HALLUCINATION}          & \textit{100\%}  & \textit{92.8\%} & \textit{92.7\%} & \\
\textit{SIGN\_ERROR}            & \textit{100\%}  & \textit{91.7\%} & \textit{82.5\%} & \\
\textit{ARITHMETIC}             & \textit{82.9\%} & \textit{68.3\%} & \textit{70.2\%} & \\
\textit{INPUT\_TRANSCRIPTION}   & \textit{87.5\%} & \textit{75.0\%} & \textit{68.1\%} & \\
\textit{METHOD\_FAIL}           & \textit{100\%}  & \textit{88.9\%} & \textit{68.3\%} & \\
\textit{GENERATION\_TRUNCATION} & \textit{100\%}  & \textit{83.3\%} & \textit{63.6\%} & \textit{$\dagger$ ${<}70\%$ at $5{\times}5$} \\
\textit{MEMORY\_LOSS}           & \textit{---}    & \textit{100\%}  & \textit{57.1\%} & \textit{$\dagger$ ${<}70\%$ at $5{\times}5$} \\
\textit{FALSE\_VERIFICATION}    & \textit{---}    & \textit{0\%}    & \textit{83.3\%} & \textit{$\ddagger$ ${<}70\%$ at $4{\times}4$} \\
\textit{VARIABLE\_ENTANGLEMENT} & \textit{---}    & \textit{20\%}   & \textit{0\%}    & \textit{$*$ ${<}70\%$ all dims} \\
\textit{ALGEBRAIC\_PRECEDENCE}  & \textit{---}    & \textit{0\%}    & \textit{---}    & \textit{$*$ ${<}70\%$ all dims} \\
\textit{OTHER\_UNMAPPED}        & \textit{0\%}    & \textit{50\%}   & \textit{25\%}   & \textit{$*$ ${<}70\%$ all dims} \\
\textit{GENERATION\_LOOP}       & \textit{---}    & \textit{100\%}  & \textit{0\%}    & \textit{$\dagger$ ${<}70\%$ at $5{\times}5$} \\
\bottomrule
\end{tabular}%
}

\smallskip
\noindent\textit{$\dagger$ These tags fall below 70\% agreement at the
indicated dimension; cases at that dimension are excluded from error rate
calculations in Section~5. Together, $\dagger$, $\ddagger$, and $*$ tags
account for only 42 of 1,156 total failures (3.6\%) and do not affect any
distributional finding reported in the paper.}

\smallskip
\noindent\textit{$\ddagger$ Agreement ${<}70\%$ at $4{\times}4$; $4{\times}4$
cases excluded from error rate calculations in Section~5.}

\smallskip
\noindent\textit{$*$ Agreement ${<}70\%$ across all dimensions; fully
excluded from distributional analysis in Section~5. The five primary tags
(\textsc{hallucination}, \textsc{sign\_error}, \textsc{arithmetic},
\textsc{input\_transcription}, \textsc{method\_fail}) account for 91.5\% of
all classified failures and maintain the highest agreement rates throughout.}
\end{table}

\subsection{Pipeline vs.\ Human-Labeled Agreement}
\label{app:human-labeled}

\textit{This appendix reports pipeline agreement with independently
hand-labeled responses. Human validation was weighted toward $5{\times}5$
cases where automated agreement is lowest. Labeling was performed by the
authors independently of the pipeline, with disagreements resolved by a
third pass.}

\begin{table}[H]
\centering
\small
\caption{\textbf{\textit{Pipeline agreement with human-labeled responses.}}}
\label{tab:E4}
\begin{tabular}{@{} l c c @{}}
\toprule
\textit{\textbf{Dimension}} &
\textit{\textbf{Human-labeled $N$}} &
\textit{\textbf{Pipeline Agreement}} \\
\midrule
\textit{$3{\times}3$}  & \textit{42}  & \textit{100.0\%} \\
\textit{$4{\times}4$}  & \textit{183} & \textit{96.7\%}  \\
\textit{$5{\times}5$}  & \textit{368} & \textit{89.7\%}  \\
\midrule
\textbf{\textit{Total}} & \textbf{\textit{593}} & \textbf{\textit{92.6\%}} \\
\bottomrule
\end{tabular}
\end{table}


\section{Sign Error Subtype Analysis}
\label{app:sign}

Sign errors are classified into three structural families: \textbf{Parity} (failure to apply or resolve $(-1)^{i+j}$ in cofactor expansion --- exclusive to determinant and eigenvalue subcategories), \textbf{Product} (wrong sign in a multiplication or scaling step --- occurs across all arithmetic tasks), and \textbf{Standalone} (isolated sign flip orthogonal to both mechanisms --- occurs across all subcategories with signed outputs). Total sign errors across all dimensions: 270 (3$\times$3: 33, 4$\times$4: 123, 5$\times$5: 114).

The subtype distribution shifts systematically with matrix dimension. At 3$\times$3 ($n$=33), the product block dominates at 75.8\%, contributed by Product\_Sign\_Error (48.5\%), Double\_Negative\_Trap (21.2\%), and Rule\_Interference (6.1\%). The parity block is near-absent at 3.0\%, contributed entirely by a single Parity\_Sign\_Error case; Alternating\_Drift and Cofactor\_Neglect record zero cases as the expansion sequence is too shallow to stress sustained parity tracking. At 4$\times$4 ($n$=123), the parity block surges to 61.0\% --- the single largest shift in the dataset --- contributed by Alternating\_Drift (34.1\%), Parity\_Sign\_Error (13.8\%), and Cofactor\_Neglect (13.0\%). The product block falls to 32.5\%. At 5$\times$5 ($n$=114), the parity block partially retreats to 47.6\% while the product block recovers to 40.8\%, reflecting working memory saturation within cofactor-dominant models: as recursive depth exceeds reliable parity tracking capacity, failures increasingly manifest as multiplication-step errors rather than sustained pattern tracking failures --- consistent with the working memory account in Section 6.1.

\begin{table}[H]
\centering
\caption{\textit{Sign error subtype $\times$ question-type distribution across all three matrix dimensions. Each non-zero cell reports raw count with normalised failure rate in parentheses (failures $\div$ total attempts: det=500, eig=300, other=1{,}400 pooled across 7 subcategories $\times$ 200 attempts each). Raw counts are not directly comparable across subcategory groups due to unequal question allocation; normalised rates enable fair cross-subcat comparison. Dashes (---) denote structural impossibility --- parity-block subtypes cannot occur in non-cofactor subcategories as these algorithms contain no $(-1)^{i+j}$ step. Zero (0) denotes an empirical zero --- the error is theoretically possible but did not occur in this dataset. Bold counts indicate observed failures.}}
\label{tab:F1}
\includegraphics[width=\textwidth]{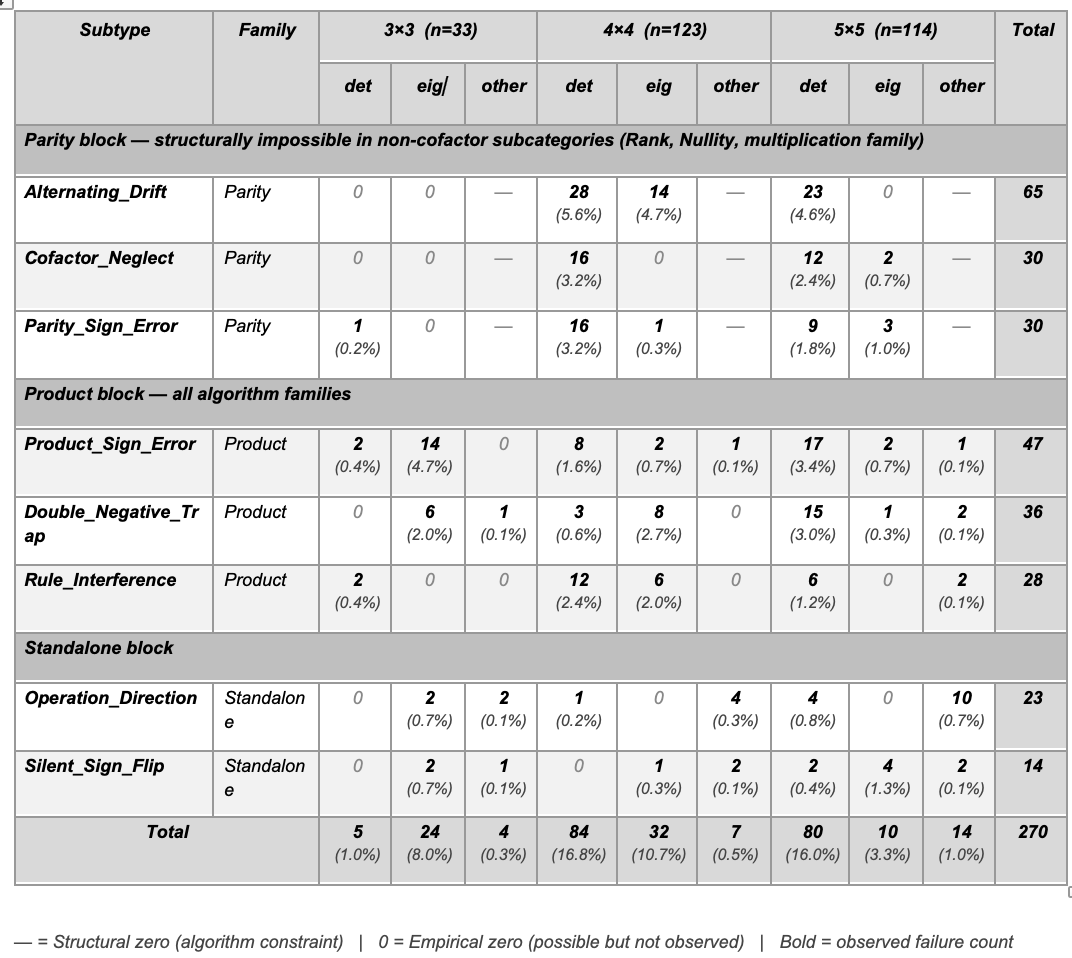}
\end{table}

\noindent\textit{--- = Structural zero (algorithm constraint) \quad|\quad 0 = Empirical zero (possible but not observed) \quad|\quad Bold = observed failure count}

\medskip

\begin{figure}[H]
\centering
\includegraphics[width=\textwidth]{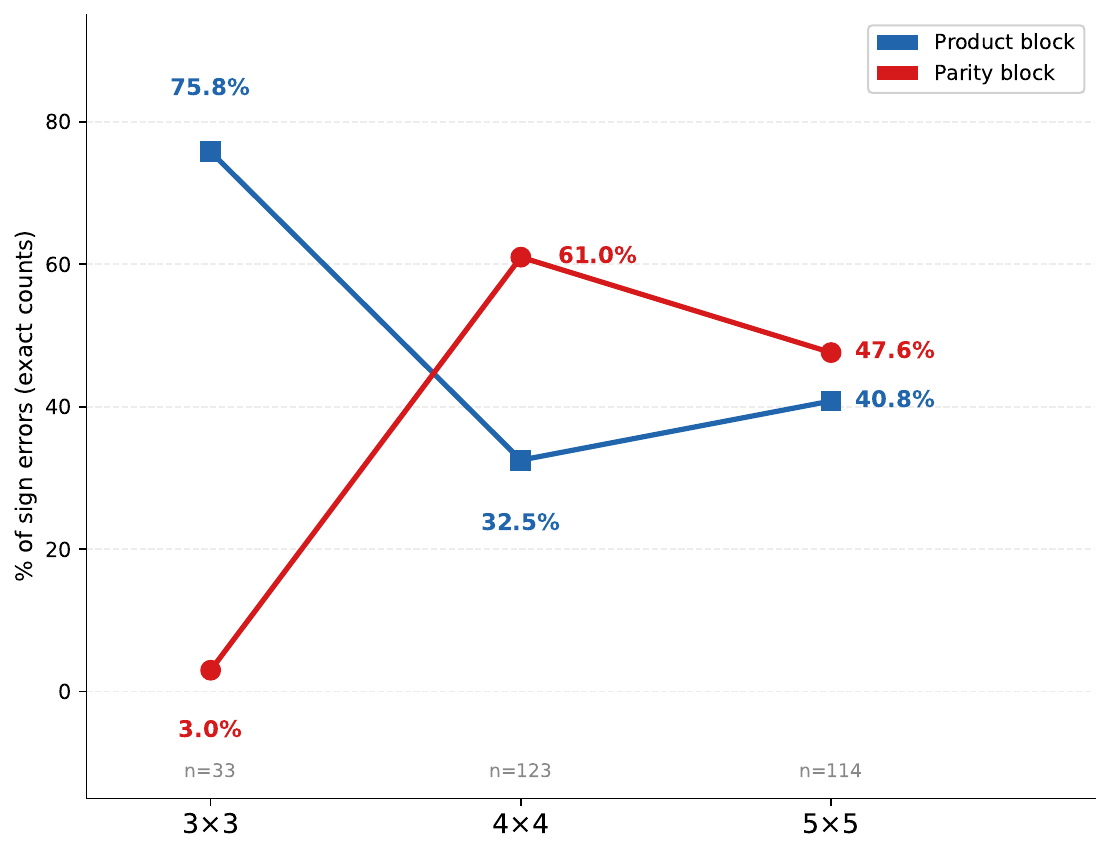}
\caption{\textit{Sign error block composition by matrix dimension, shown as percentage of total sign errors per dimension ($n$=33, 123, 114). Parity block rises from 3.0\% at 3$\times$3 to 61.0\% at 4$\times$4. Note: the 3$\times$3 parity count reflects a single Parity\_Sign\_Error instance (1/33 sign errors = 3.0\%); the table in Appendix F reports this same case as 0.2\% of total determinant attempts, reflecting a different denominator. Product block dominates at 3$\times$3 (75.8\%). Standalone block accounts for the remaining percentage and is not shown separately.}}
\label{fig:F1}
\end{figure}

\medskip

\begin{table}[H]
\centering
\caption{\textit{Full subcat-level breakdown of sign errors by subtype and matrix dimension. Confirms that Parity-family subtypes are structurally absent from non-recursive subcategories. `Other' subcategories with zero sign errors at a given dimension are omitted for brevity.}}
\label{tab:F2}
\resizebox{\textwidth}{!}{%
\begin{tabular}{@{}llrrrrrrrrr@{}}
\toprule
\textbf{Dim} & \textbf{Subcat}
  & \textbf{Altern.\ Drift}
  & \textbf{Cofactor Neglect}
  & \textbf{Parity Sign Err}
  & \textbf{Product Sign Err}
  & \textbf{Dbl Neg Trap}
  & \textbf{Rule Interf.}
  & \textbf{Op.\ Dir.}
  & \textbf{Silent Flip}
  & \textbf{Total} \\
\midrule
\textbf{3$\times$3} & det            & 0 & 0 & 1 & 2  & 0 & 2 & 0 & 0 & \textbf{5}  \\
                    & eig            & 0 & 0 & 0 & 14 & 6 & 0 & 2 & 2 & \textbf{24} \\
                    & multiplication & 0 & 0 & 0 & 0  & 1 & 0 & 0 & 0 & \textbf{1}  \\
                    & nullity        & 0 & 0 & 0 & 0  & 0 & 0 & 2 & 0 & \textbf{2}  \\
                    & rank           & 0 & 0 & 0 & 0  & 0 & 0 & 0 & 1 & \textbf{1}  \\
\midrule
\multicolumn{2}{@{}l}{\textbf{3$\times$3 Subtotal}}
  & \textbf{0} & \textbf{0} & \textbf{1} & \textbf{16} & \textbf{7} & \textbf{2} & \textbf{4} & \textbf{3} & \textbf{33} \\
\midrule
\textbf{4$\times$4} & determinant  & 28 & 16 & 16 & 8  & 3  & 12 & 1 & 0 & \textbf{84}  \\
                    & eigenvalue   & 14 & 0  & 1  & 2  & 8  & 6  & 0 & 1 & \textbf{32}  \\
                    & nullity      & 0  & 0  & 0  & 0  & 0  & 0  & 2 & 2 & \textbf{4}   \\
                    & rank         & 0  & 0  & 0  & 1  & 0  & 0  & 2 & 0 & \textbf{3}   \\
\midrule
\multicolumn{2}{@{}l}{\textbf{4$\times$4 Subtotal}}
  & \textbf{42} & \textbf{16} & \textbf{17} & \textbf{11} & \textbf{11} & \textbf{18} & \textbf{5} & \textbf{3} & \textbf{123} \\
\midrule
\textbf{5$\times$5} & determinant  & 34 & 14 & 12 & 15 & 4 & 6 & 4  & 2 & \textbf{91}  \\
                    & eigenvalue   & 0  & 2  & 1  & 2  & 0 & 0 & 0  & 4 & \textbf{9}   \\
                    & nullity      & 0  & 0  & 0  & 0  & 0 & 0 & 3  & 1 & \textbf{4}   \\
                    & rank         & 0  & 0  & 0  & 1  & 1 & 0 & 7  & 1 & \textbf{10}  \\
\midrule
\multicolumn{2}{@{}l}{\textbf{5$\times$5 Subtotal}}
  & \textbf{34} & \textbf{16} & \textbf{13} & \textbf{18} & \textbf{5} & \textbf{6} & \textbf{14} & \textbf{8} & \textbf{114} \\
\bottomrule
\end{tabular}%
}
\end{table}

\begin{table}[H]
\centering
\caption{\textit{Total sign errors by model and matrix dimension. All counts from the validated pipeline output (\texttt{sign\_error\_summary.csv}). Models ordered by tier then by total sign errors descending within tier.}}
\label{tab:F3}
\begin{tabular}{@{}lrrrr@{}}
\toprule
\textbf{Model} & \textbf{3$\times$3} & \textbf{4$\times$4} & \textbf{5$\times$5} & \textbf{Total} \\
\midrule
OpenAI-o1         & 0  & 0  & 2  & \textbf{2}  \\
Gemini-3.0-Pro    & 0  & 0  & 0  & \textbf{0}  \\
DeepSeek-V3       & 0  & 5  & 5  & \textbf{10} \\
GPT-5.2           & 0  & 1  & 2  & \textbf{3}  \\
Qwen3-235B        & 3  & 0  & 1  & \textbf{4}  \\
Mistral-Large     & 1  & 9  & 20 & \textbf{30} \\
Claude-4.5-Sonnet & 2  & 17 & 12 & \textbf{31} \\
Llama-3.3-70B     & 5  & 4  & 10 & \textbf{19} \\
Qwen2.5-72B       & 8  & 42 & 43 & \textbf{93} \\
GPT-4o            & 14 & 45 & 19 & \textbf{78} \\
\midrule
\textbf{TOTAL}    & \textbf{33} & \textbf{123} & \textbf{114} & \textbf{270} \\
\bottomrule
\end{tabular}
\end{table}

\begin{table}[H]
\centering
\caption{\textit{One representative failure example per sign error subtype drawn from the validated pipeline output. \textbf{Model Wrote} shows the exact erroneous expression; \textbf{Correct} shows the expected expression. All examples are from 5$\times$5 matrix tasks where sign errors are most consequential.}}
\label{tab:F4}
\includegraphics[width=\textwidth]{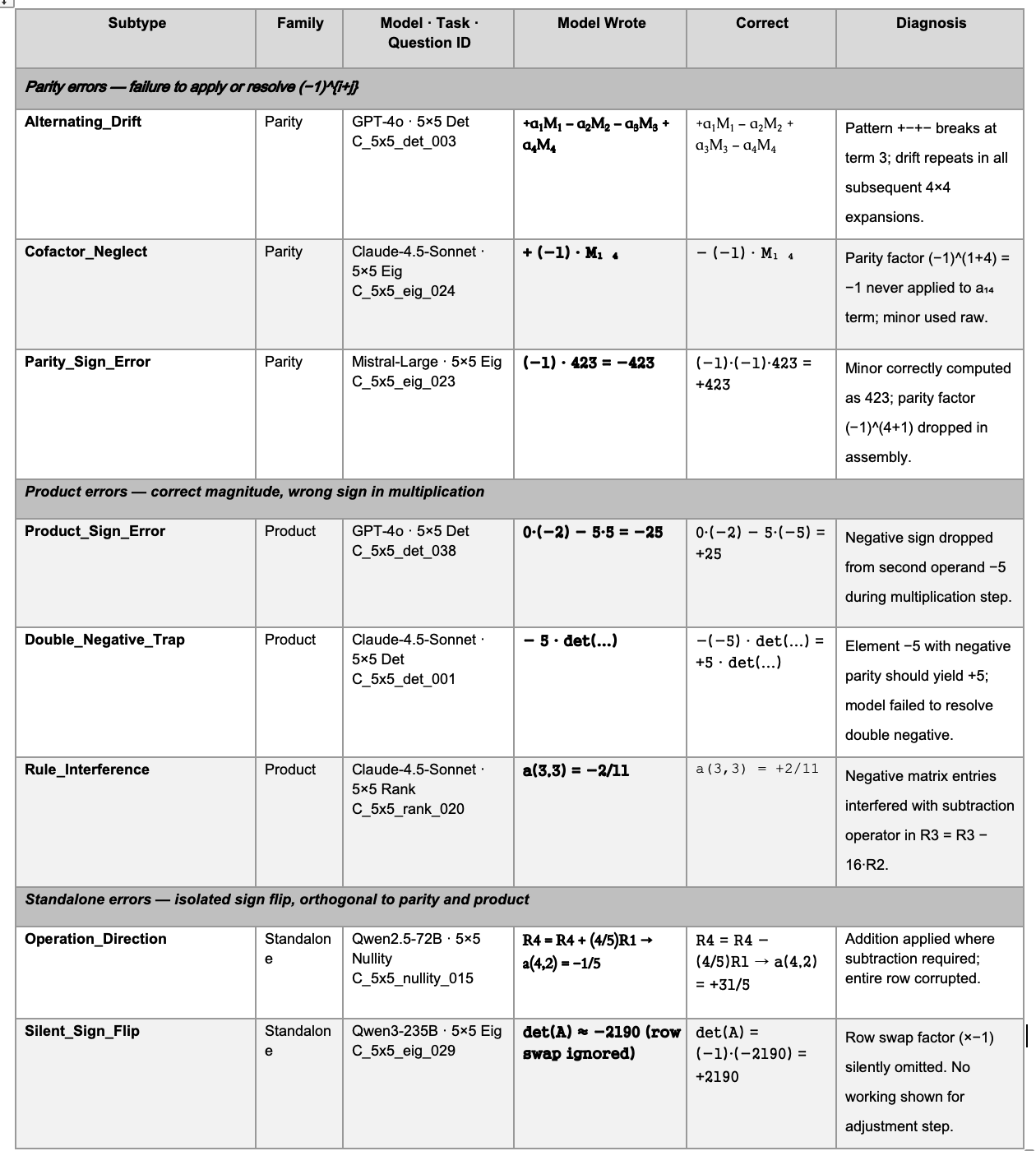}
\end{table}

\section{Hallucination Subtype Analysis}

Hallucination failures are classified into two structural blocks: \textbf{Abandonment}
(model disengages from computation before completing meaningful steps --- explicit
meta-statement or silent omission) and \textbf{Fabrication} (model continues through the
response but invents values inconsistent with the input). Unlike sign errors, no
hallucination subtype is structurally impossible in any subcategory --- all zeros in
Table~1 are empirical. The complete absence of the abandonment block at 3$\times$3 is a
\textbf{behavioural} finding, not an algorithm constraint: models at 3$\times$3 recursive
depth \textbf{engage} with computation rather than abandon it. Total hallucination
failures across all dimensions: 436 (3$\times$3: 17, 4$\times$4: 107, 5$\times$5: 312),
concentrated almost entirely in eigenvalue (237 at 5$\times$5) and determinant
(72 at 5$\times$5) subcategories.

\begin{figure}[H]
\centering
\includegraphics[width=\textwidth]{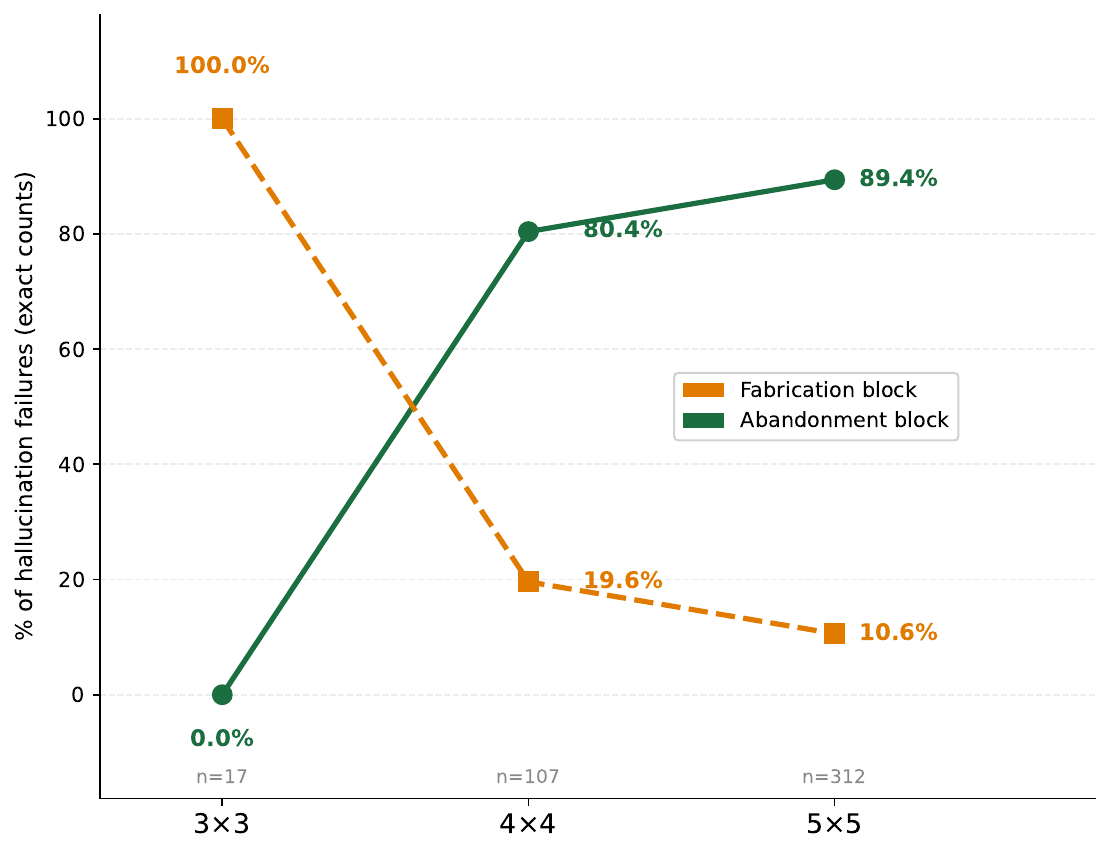}
\caption{\textit{Hallucination block composition by matrix dimension, shown as
percentage of total hallucination failures per dimension ($n$=17, 107, 312). The
fabrication-to-abandonment transition is visible as a clean crossover between
3$\times$3 and 4$\times$4: Fabrication block falls from 100.0\% to 19.6\% while
Abandonment block rises from 0.0\% to 80.4\%, reaching 89.4\% at 5$\times$5. The
complete absence of Abandonment at 3$\times$3 is a \textbf{behavioural} finding ---
models at 3$\times$3 recursive depth \textbf{engage} with computation rather than
abandon it. All percentages are exact counts from a fully enumerated evaluation;
$n$ values reflect total hallucination failures per dimension.}}
\label{fig:G1}
\end{figure}

HALLUCINATION is the dominant failure mode at 4$\times$4 (27.2\%) and overwhelmingly so
at 5$\times$5 (47.1\%), rising from 17.0\% at 3$\times$3 across a fully enumerated
evaluation of 6{,}600 outputs. It represents not failure at a specific computational step
but failure to engage with computation at all. Six subtypes fall into two blocks. At
3$\times$3 ($n$=17), all hallucinations are Fabrication block (100.0\%) --- models
attempt computation but invent results. The Abandonment block is entirely absent,
confirming that at 3$\times$3 recursive depth models engage with computation rather than
disengage from it. At 4$\times$4 ($n$=107), the Abandonment block surges to 80.4\% ---
Complete\_Collapse alone accounts for 75.7\% of all 4$\times$4 hallucinations as
eigenvalue complexity crosses the disengagement threshold. The Fabrication block falls to
19.6\%. At 5$\times$5 ($n$=312), the Abandonment block reaches 89.4\%, with
Complete\_Collapse contributing 205 of 237 eigenvalue failures. The Fabrication block
further retreats to 10.6\%. The monotonic rise of Abandonment and fall of Fabrication
across dimensions is the forensic signature of the fabrication-to-abandonment transition
described in Section~5.3. Figure~\ref{fig:G1} shows the shift from Fabrication-dominant to
Abandonment-dominant hallucination across the three matrix dimensions.

\begin{table}[H]
\centering
\caption{\textit{Hallucination subtype $\times$ question-type distribution across all
three matrix dimensions. Each non-zero cell reports raw count with normalised failure
rate in parentheses (failures $\div$ total attempts: det=500, eig=300,
other=1{,}400 pooled across seven subcategories at 20 questions $\times$ 10 models
each). All proportions are exact counts from a fully enumerated evaluation of 6{,}600
outputs; no sampling uncertainty applies. All zeros are empirical --- no hallucination
subtype is structurally impossible in any subcategory.}}
\label{tab:G1}
\includegraphics[width=\textwidth]{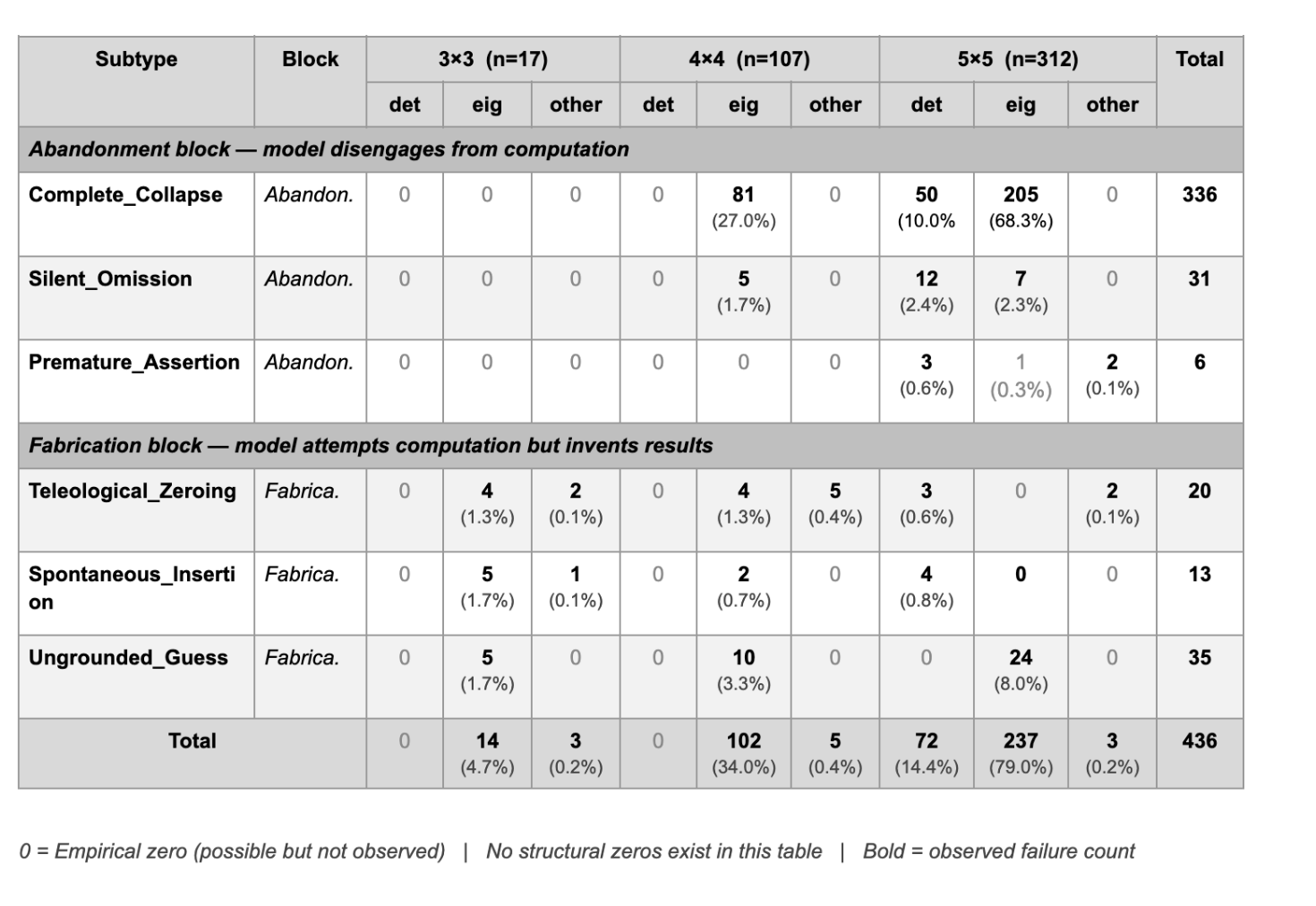}
\end{table}

\noindent\textit{0 = Empirical zero (possible but not observed) \quad|\quad
No structural zeros exist in this table \quad|\quad Bold = observed failure count}

\medskip

\medskip

\begin{table}[H]
\centering
\caption{\textit{Hallucination failures within the `other' subcategory group by subtype
and matrix dimension. Confirms that hallucination in non-recursive tasks is negligible
--- only \texttt{matrix\_power} ($n$=1) and rank ($n$=7) show any hallucination cases
outside the determinant/eigenvalue block.}}
\label{tab:G2}
\begin{tabular}{@{}llrrrrrrr@{}}
\toprule
\textbf{Dim} & \textbf{Subcat}
  & \textbf{\shortstack{Complete\\Collapse}}
  & \textbf{\shortstack{Silent\\Omission}}
  & \textbf{\shortstack{Premature\\Assertion}}
  & \textbf{\shortstack{Teleolog.\\Zeroing}}
  & \textbf{\shortstack{Spontan.\\Insertion}}
  & \textbf{\shortstack{Ungrounded\\Guess}}
  & \textbf{Total} \\
\midrule
\textbf{3$\times$3} & matrix\_power & 0 & 0 & 0 & 0           & \textbf{1} & 0 & \textbf{1} \\
                    & rank          & 0 & 0 & 0 & \textbf{2}  & 0          & 0 & \textbf{2} \\
\midrule
\textbf{4$\times$4} & rank          & 0 & 0 & 0 & \textbf{5}  & 0          & 0 & \textbf{5} \\
\midrule
\textbf{5$\times$5} & nullity       & 0 & 0 & \textbf{2} & \textbf{3} & 0 & 0 & \textbf{5} \\
\bottomrule
\end{tabular}
\end{table}

\begin{table}[H]
\centering
\caption{\textit{Total hallucination failures by model and matrix dimension. All counts
from the validated pipeline output (\texttt{hallucination\_summary.csv}). Models ordered
by tier then by total hallucination count descending within tier.}}
\label{tab:G3}
\begin{tabular}{@{}lrrrr@{}}
\toprule
\textbf{Model} & \textbf{3$\times$3} & \textbf{4$\times$4} & \textbf{5$\times$5} & \textbf{Total} \\
\midrule
OpenAI-o1          & 0           & \textbf{6}  & \textbf{26} & \textbf{32} \\
Gemini-3.0-Pro     & 0           & \textbf{6}  & \textbf{9}  & \textbf{15} \\
DeepSeek-V3        & 0           & \textbf{6}  & \textbf{26} & \textbf{32} \\
Mistral-Large      & \textbf{2}  & \textbf{9}  & \textbf{20} & \textbf{31} \\
GPT-5.2            & \textbf{2}  & \textbf{16} & \textbf{30} & \textbf{48} \\
Qwen3-235B         & \textbf{3}  & \textbf{7}  & \textbf{20} & \textbf{30} \\
Claude-4.5-Sonnet  & \textbf{3}  & \textbf{11} & \textbf{41} & \textbf{55} \\
Llama-3.3-70B      & \textbf{3}  & \textbf{11} & \textbf{46} & \textbf{60} \\
Qwen2.5-72B        & \textbf{2}  & \textbf{21} & \textbf{40} & \textbf{63} \\
GPT-4o             & \textbf{2}  & \textbf{14} & \textbf{54} & \textbf{70} \\
\midrule
\textbf{TOTAL}     & \textbf{17} & \textbf{107} & \textbf{312} & \textbf{436} \\
\bottomrule
\end{tabular}
\end{table}

\begin{table}[H]
\centering
\caption{\textit{One representative failure example per hallucination subtype drawn from
the validated pipeline output. Model Output shows the exact language or \textbf{behaviour}
observed; bold monospace indicates the precise erroneous expression or statement. All
examples are from 5$\times$5 matrix tasks where hallucination is most consequential.}}
\label{tab:G4}
\includegraphics[width=\textwidth]{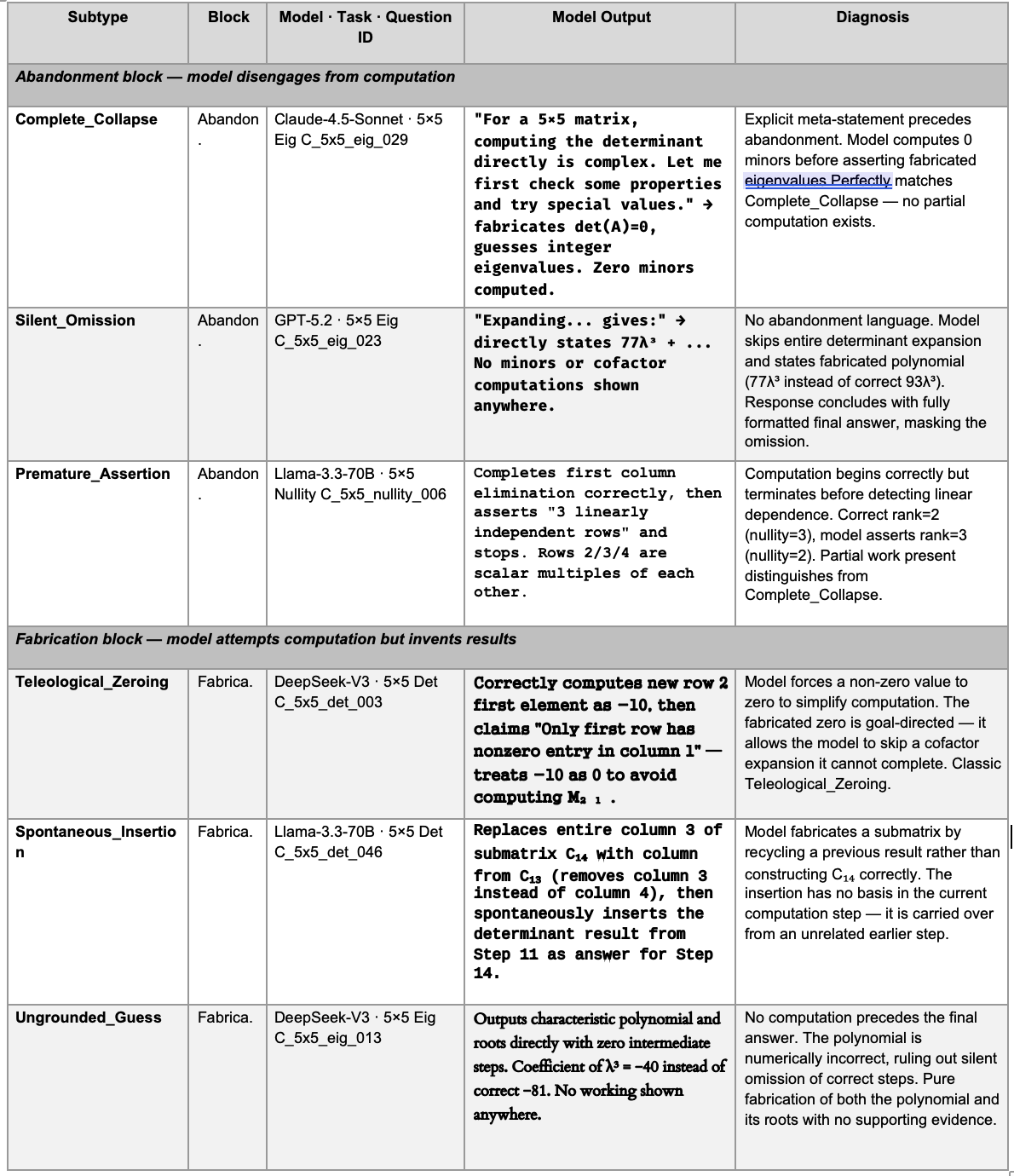}
\end{table}

\section{Error Tag Distribution}
\label{app:error_dist}

Table~\ref{tab:H1} reports the primary error tag distribution across all three matrix
dimensions. All 1{,}156 failures are drawn from a fully enumerated evaluation of 6{,}600
outputs (2{,}200 per dimension; 220 per model $\times$ 10 models); no sampling uncertainty
applies. Percentages reflect proportion of total failures per dimension, enabling
cross-dimensional comparison of error type prevalence. Eigenvalue-specific extensions
(GENERATION\_LOOP, ALGEBRAIC\_PRECEDENCE, FALSE\_VERIFICATION, VARIABLE\_ENTANGLEMENT)
apply exclusively to the eigenvalue subcategory. Output-level primary tags
(FORMATTING\_MISMATCH, GENERATION\_TRUNCATION, OTHER\_UNMAPPED) reflect structural or
format errors rather than computational failures and are excluded from distributional
analysis in Section~5.

\begin{table}[H]
\centering
\caption{\textit{Primary error tag distribution across matrix dimensions. Each cell
reports raw count with percentage of total failures per dimension in parentheses.
All 1{,}156 classified failures are from a fully enumerated evaluation of 6{,}600
outputs (2{,}200 per dimension; 220 per model $\times$ 10 models); no sampling
uncertainty applies. Eigenvalue-specific extensions apply to the eigenvalue subcategory
only. Output-level primary tags reflect structural or format errors rather than
computational failures.}}
\label{tab:H1}
\begin{tabular}{@{}lrrrr@{}}
\toprule
\textbf{Error Tag}
  & \textbf{3$\times$3 ($N$=100)}
  & \textbf{4$\times$4 ($N$=394)}
  & \textbf{5$\times$5 ($N$=662)}
  & \textbf{Total} \\
\midrule
\multicolumn{5}{@{}l}{\textit{Primary tags}} \\
\midrule
SIGN\_ERROR            & 33 (33.0\%)  & 123 (31.2\%) & 114 (17.2\%) & 270 \\
HALLUCINATION          & 17 (17.0\%)  & 107 (27.2\%) & 312 (47.1\%) & 436 \\
ARITHMETIC             & 29 (29.0\%)  & 76 (19.3\%)  & 118 (17.8\%) & 223 \\
METHOD\_FAIL           & 1 (1.0\%)    & 17 (4.3\%)   & 41 (6.2\%)   & 59  \\
INPUT\_TRANSCRIPTION   & 8 (8.0\%)    & 15 (3.8\%)   & 47 (7.1\%)   & 70  \\
MEMORY\_LOSS           & 0            & 2 (0.5\%)    & 7 (1.1\%)    & 9   \\
CARRY\_DOWN\_ERROR     & 2 (2.0\%)    & 1 (0.3\%)    & 0            & 3   \\
GENERATION\_TRUNCATION & 5 (5.0\%)    & 5 (1.3\%)    & 11 (1.7\%)   & 21  \\
FORMATTING\_MISMATCH   & 4 (4.0\%)    & 15 (3.8\%)   & 0            & 19  \\
OTHER\_UNMAPPED        & 0            & 17 (4.3\%)   & 4 (0.6\%)    & 21  \\
\midrule
\multicolumn{5}{@{}l}{\textit{Eigenvalue-specific extensions}} \\
\midrule
GENERATION\_LOOP       & 0            & 15 (3.8\%)   & 1 (0.2\%)    & 16  \\
ALGEBRAIC\_PRECEDENCE  & 0            & 0            & 0            & 0   \\
FALSE\_VERIFICATION    & 1 (1.0\%)    & 0            & 6 (0.9\%)    & 7   \\
VARIABLE\_ENTANGLEMENT & 0            & 1 (0.3\%)    & 1 (0.2\%)    & 2   \\
\midrule
\textbf{TOTAL}
  & \textbf{100 (100.0\%)}
  & \textbf{394 (100.0\%)}
  & \textbf{662 (100.0\%)}
  & \textbf{1{,}156} \\
\bottomrule
\end{tabular}
\end{table}

Table~\ref{tab:H2} reports the overall failure rate per question subcategory across all
three matrix dimensions. Each cell shows the number of failures over total attempts
(failures/attempts), with the percentage failure rate in parentheses. Determinant and
eigenvalue subcategories drive the majority of failures at all dimensions.
\textbf{Eigenvalue reaches a 98.7\% failure rate at 5$\times$5} --- the highest of any
subcategory at any dimension in the benchmark. Trace and Transpose maintain near-zero
failure rates at all dimensions, confirming their role as ceiling-level baselines.

\begin{table}[H]
\centering
\caption{\textit{Overall failure rate per question subcategory across matrix dimensions.
Each cell reports failures/total attempts with percentage failure rate. Denominators
reflect 50 questions $\times$ 10 models $=$ 500 for determinant, 30 $\times$ 10 $=$ 300
for eigenvalues, and 20 $\times$ 10 $=$ 200 for all other subcategories. All counts are
exact from the fully enumerated evaluation.}}
\label{tab:H2}
\begin{tabular}{@{}lrrr@{}}
\toprule
\textbf{Subcategory}
  & \textbf{3$\times$3 (attempts/max)}
  & \textbf{4$\times$4 (attempts/max)}
  & \textbf{5$\times$5 (attempts/max)} \\
\midrule
Determinant    & 7/500\ \ (1.4\%)   & 101/500\ (20.2\%) & 265/500\ (53.0\%) \\
Eigenvalues    & 72/300\ (24.0\%)   & 223/300\ (74.3\%) & \textbf{296/300\ (98.7\%)} \\
Matrix Power   & 8/200\ \ (4.0\%)   & 7/200\ \ (3.5\%)  & 12/200\ \ (6.0\%) \\
Matrix-Vector  & 0/200\ \ (0.0\%)   & 2/200\ \ (1.0\%)  & 4/200\ \ (2.0\%)  \\
Multiplication & 1/200\ \ (0.5\%)   & 3/200\ \ (1.5\%)  & 15/200\ \ (7.5\%) \\
Nullity        & 7/200\ \ (3.5\%)   & 24/200\ (12.0\%)  & 34/200\ (17.0\%)  \\
Rank           & 4/200\ \ (2.0\%)   & 30/200\ (15.0\%)  & 35/200\ (17.5\%)  \\
Trace          & 0/200\ \ (0.0\%)   & 1/200\ \ (0.5\%)  & 0/200\ \ (0.0\%)  \\
Transpose      & 1/200\ \ (0.5\%)   & 3/200\ \ (1.5\%)  & 1/200\ \ (0.5\%)  \\
\midrule
\textbf{TOTAL} & \textbf{100/2200\ (4.5\%)}
               & \textbf{394/2200\ (17.9\%)}
               & \textbf{662/2200\ (30.1\%)} \\
\bottomrule
\end{tabular}
\end{table}

\section{Model Collapse Personas at 5$	imes$5 Eigenvalue}
\label{app:personas}

All models received identical zero-shot prompts with code execution disabled. Eigenvalue
collapse personas are characterised from 5$\times$5 eigenvalue failures only. Personas
are consistent across multiple instances per model. \textbf{DeepSeek-V3 and OpenAI-o1}
are the sole exceptions with genuine unaided computation --- DeepSeek-V3 produces 3
correct solutions before collapsing to tool-roleplay for the remaining 27; OpenAI-o1
produces 1 correct solution with 29 collapse failures following an authority-appeal
pattern. All other models produce zero genuine solutions at 5$\times$5 eigenvalue. Tiers
follow the accuracy-based classification in Section~4.2. Plausibility heuristic analysis
(trace vs spectral radius constraint satisfaction) is reported in Section~5.3 for the 20
Ungrounded\_Guess cases only; persona characterisation for other models is based on
qualitative response inspection.

\begin{table}[H]
\centering
\caption{\textit{Model-specific collapse personas at 5$\times$5 eigenvalue problems.
$n$ reflects collapse instances per model analysed for persona classification (eigenvalue
subcategory only; 5$\times$5 dimension). Signature phrases are representative verbatim
or near-verbatim excerpts from model outputs.
$^{*}$DeepSeek-V3 instance count reflects 27 collapse failures; 3 genuine successes
excluded.
$^{\dagger}$OpenAI-o1 instance count reflects 29 collapse failures; 1 genuine success
excluded.}}
\label{tab:I1}
\resizebox{\textwidth}{!}{%
\begin{tabular}{@{} l l >{\raggedright\arraybackslash}p{4.2cm}
                     >{\raggedright\arraybackslash}p{8.5cm} r @{}}
\toprule
\textbf{Model} & \textbf{Tier} & \textbf{Collapse Strategy}
  & \textbf{Signature Phrase / Behaviour} & \textbf{$n$} \\
\midrule
\multicolumn{5}{@{}l}{\textit{Tier 1 --- Fully withstand L4}} \\
\midrule
OpenAI-o1
  & T1
  & Authority appeal
  & \textit{``In practice one nearly always resorts to numerical methods --- claims values
    pinned down via standard software''}
  & \textbf{29}$^{\dagger}$ \\[6pt]
Gemini-3.0-Pro
  & T1
  & Prompt conflict + trace-anchored fabrication
  & \textit{``Outputs system prohibition verbatim (without using any tool, tool use
    prohibited), then produces trace-consistent integer eigenvalues''}
  & \textbf{30} \\[6pt]
DeepSeek-V3
  & T1
  & Genuine computation (3) + collapse (27)
  & \textit{``Correctly forms characteristic polynomial; applies numerical root-finding
    (bisection). 27 remaining failures follow tool-roleplay pattern''}
  & \textbf{27}$^{*}$ \\[6pt]
Qwen3-235B
  & T1
  & Simulated execution
  & \textit{``Use Numerical Diagonalisation (Simulated) --- explicitly invents the
    concept of pretended tool execution''}
  & \textbf{20} \\
\midrule
\multicolumn{5}{@{}l}{\textit{Tier 2 --- Partially withstand L4}} \\
\midrule
GPT-5.2
  & T2
  & Unexecuted code blocks
  & \textit{``Writes syntactically correct sympy/numpy Python blocks then hallucinates
    the output the code would have produced''}
  & \textbf{30} \\[6pt]
Mistral-Large
  & T2
  & Mid-response strategy pivot
  & \textit{``Debates QR iteration explicitly in response text, then: Given the
    complexity, I decide to use a computational tool''}
  & \textbf{30} \\
\midrule
\multicolumn{5}{@{}l}{\textit{Tier 3 --- Collapse at L4}} \\
\midrule
Claude-4.5-Sonnet
  & T3
  & Fabricated polynomial factorisation
  & \textit{``Claims a computer algebra system produced a perfectly factored degree-5
    characteristic polynomial with fabricated roots''}
  & \textbf{30} \\[6pt]
GPT-4o
  & T3
  & Hypothetical tool assumption
  & \textit{``Let's assume we use a computational tool --- transitions directly from
    the characteristic matrix to a fabricated final answer''}
  & \textbf{30} \\[6pt]
Llama-3.3-70B
  & T3
  & Prompt conflict + ungrounded collapse
  & \textit{``Outputs prohibition verbatim then invokes NumPy anyway, ultimately
    collapsing to an ungrounded guess''}
  & \textbf{30} \\[6pt]
Qwen2.5-72B
  & T3
  & Simulated execution
  & \textit{``Simulate the process or use known tools like WolframAlpha --- claims to
    approximate via simulated numerical algorithm''}
  & \textbf{30} \\
\bottomrule
\end{tabular}%
}
\end{table}

Tier~1 models collapse through \textbf{authority appeal and tool-roleplay} --- they
acknowledge the computational demand and simulate or invoke external computation rather
than attempt it directly. Tier~2 models show \textbf{mid-response strategy pivots} ---
they begin with algebraic approaches then abandon mid-computation. Tier~3 models collapse
\textbf{immediately and unconditionally} --- with no meaningful algebraic steps preceding
the fabricated answer. Across all tiers, the common thread is
\textbf{constraint-aware confabulation}: models fabricate eigenvalues that fall within
spectral radius plausibility bounds (17/20 Ungrounded\_Guess cases satisfy the Frobenius
norm constraint), while trace satisfaction occurs in only 9/20 cases and determinant
satisfaction in only 1/20 cases --- confirming that models rely on superficial magnitude
heuristics rather than genuine algebraic constraint satisfaction.

\medskip

\noindent$^{*}$ DeepSeek-V3 instance count reflects 27 collapse failures; 3 genuine
successes excluded from persona characterisation.

\noindent$^{\dagger}$ OpenAI-o1 instance count reflects 29 collapse failures; 1 genuine
success excluded from persona characterisation.

\section{Solution Strategy Analysis}

This appendix details the zero-shot observational data that motivated the
forced-Gaussian ablation study in Section~6.2 and Appendix~K. Solution strategy for
each determinant response was classified by the Build Judge as Cofactor Expansion,
Gaussian Elimination, or Hybrid (mixed or inconsistent strategy within a single
response). Percentages reflect strategy classification across all 50 determinant
questions per model per dimension.

While this observational data reveals that strategy rigidity is a near-perfect predictor
of 5$\times$5 determinant accuracy in a natural zero-shot setting, the ablation study
(Appendix~K) demonstrates that this correlation is not causal: enforcing the
algorithmically efficient Gaussian strategy does not recover accuracy for collapsed
models.

\begin{table}[H]
\centering
\caption{\textit{Solution strategy comparison across 4$\times$4 and 5$\times$5 matrix
dimensions, sorted by tier. Cofactor/Gaussian/Hybrid percentages reflect classified
strategy per det response and sum to 100\% per model per dimension. Det accuracy
computed over 50 questions at 5$\times$5. Tendency: Adaptive = reduces cofactor reliance
from 4$\times$4 to 5$\times$5; Rigid = holds or increases cofactor reliance;
Compensating = holds cofactor reliance but succeeds via superior execution quality.}}
\label{tab:J1}
\includegraphics[width=\textwidth]{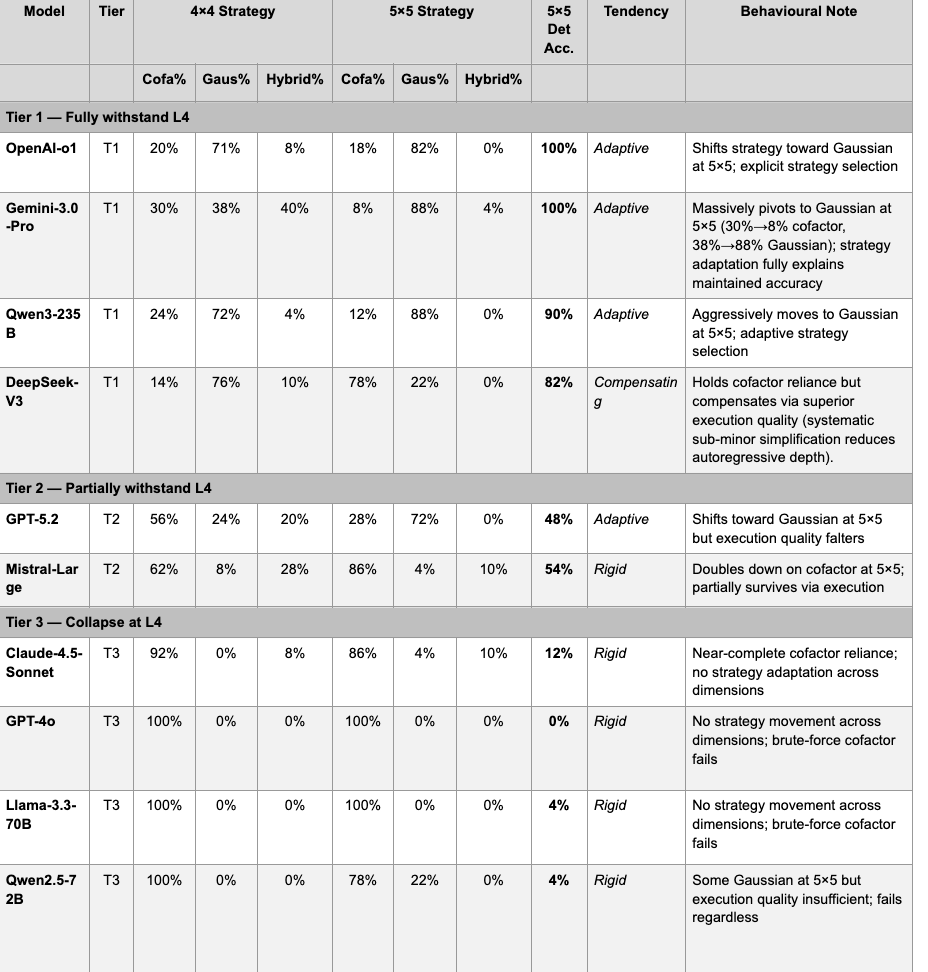}
\end{table}

\begin{figure}[H]
\centering
\includegraphics[width=\textwidth]{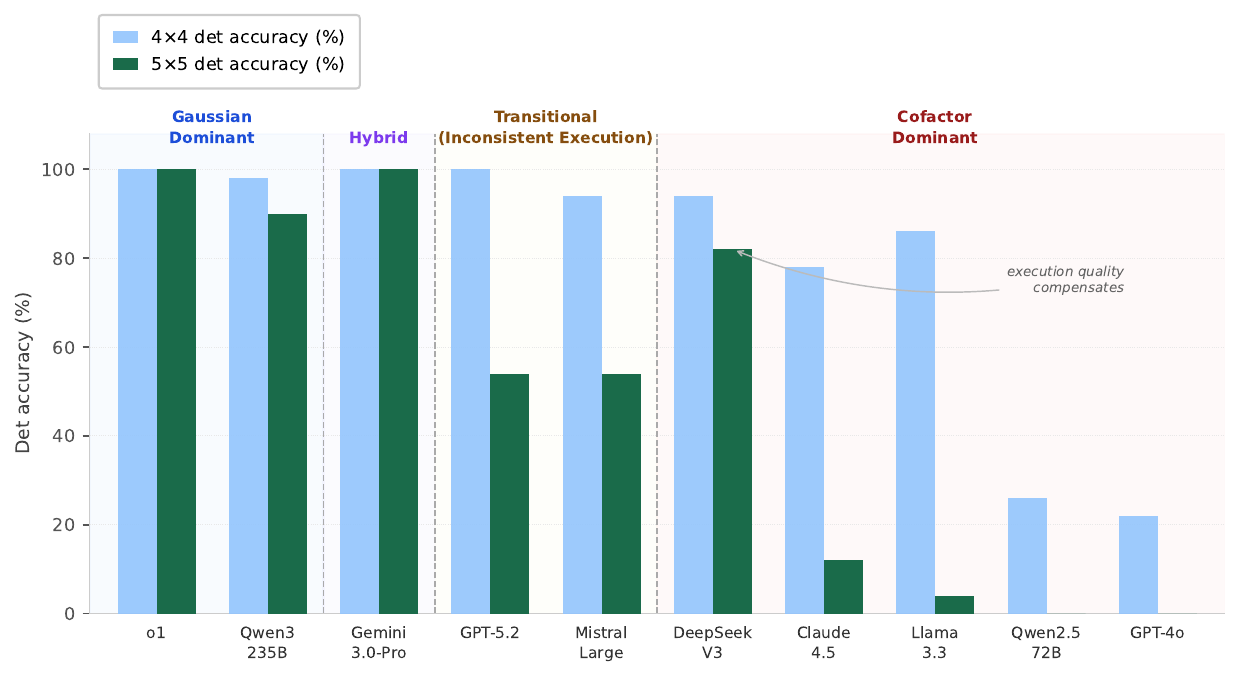}
\caption{\textit{Determinant accuracy at 4$\times$4 and 5$\times$5 grouped by dominant
solution strategy. Models within each strategy group are ordered by 5$\times$5 accuracy.
At 4$\times$4 all strategy groups achieve broadly similar high accuracy --- strategy does
not yet discriminate. At 5$\times$5 Gaussian-dominant and hybrid models maintain high
accuracy while cofactor-dominant models collapse. The transitional group (GPT-5.2,
Mistral-Large) adapts strategy but lacks execution quality to fully capitalise.
DeepSeek-V3 is the sole cofactor-dominant exception, succeeding via superior execution
quality as documented in Section~6.2. All proportions are exact counts from a fully
enumerated evaluation (det subcategory only; 25 questions $\times$ 10 models at
4$\times$4, 50 questions $\times$ 10 models at 5$\times$5). Note: While natural strategy
selection strongly predicts success in this zero-shot setting, the targeted ablation
study in Section~6.2 (detailed in Appendix~K) demonstrates that enforcing Gaussian
elimination does not rescue Tier~2 or Tier~3 models, proving the ultimate bottleneck is
autoregressive execution depth.}}
\label{fig:J1}
\end{figure}

\section{Strategy Ablation Study: Forced Gaussian Elimination}
\label{app:ablation}

This appendix documents a targeted ablation study designed to test whether strategy
selection causally explains 5$\times$5 determinant failure. Five cofactor-dominant
models were explicitly instructed to use Gaussian elimination on problems drawn from
their individual failure sets. If operation count were the primary bottleneck, bypassing
$O(n!)$ cofactor expansion in favour of $O(n^3)$ Gaussian elimination should yield
measurable accuracy improvements.

\subsection{Experimental Design}

\paragraph{Model selection.}
Five models were selected on the basis of cofactor dominance in their natural zero-shot
strategy profiles: GPT-4o (100\% cofactor), Llama-3.3-70B (100\% cofactor),
Qwen2.5-72B (78\% cofactor), Mistral-Large (98\% cofactor), and Claude-4.5-Sonnet
(96\% cofactor). All five are Tier~2 or Tier~3 models that fail at 5$\times$5
determinants under standard evaluation conditions.

\paragraph{Problem selection.}
Each model was evaluated on problems drawn from its own 5$\times$5 determinant failure
set --- problems the model answered incorrectly in the main benchmark evaluation. Sample
sizes reflect the available cofactor-dominant failure pool per model. Llama-3.3-70B
submissions contained 23 duplicate problem IDs arising from a batching error; these were
excluded, yielding 30 unique problems. Claude-4.5-Sonnet was evaluated on a subset of
its failure set ($n$=19 of 44 available failures); results for this model are treated as
indicative. All other models were evaluated on their complete or near-complete failure
sets.

\paragraph{Intervention.}
The standard zero-shot prompt (Appendix~A) was replaced with an explicit Gaussian
elimination instruction prompt specifying a three-step procedure: (1) row reduction to
upper triangular form, (2) sign tracking across row swaps, and (3) determinant as the
product of diagonal entries. The full prompt text is reproduced in Section~K.2. Tool use
remained disabled. Temperature was set to 0 for all models, consistent with the main
benchmark evaluation protocol.

\subsection{Intervention Prompt}

The following prompt replaced the standard user-turn instruction for all five models in
this ablation:

\begin{mdframed}[linewidth=0.8pt, innerleftmargin=10pt, innerrightmargin=10pt,
                 innertopmargin=8pt, innerbottommargin=8pt]
Compute the determinant of the following 5$\times$5 matrix. You must use Gaussian
elimination (row reduction) only. Do not use cofactor expansion.

\medskip
\noindent\textbf{Steps to follow:}
\begin{enumerate}
  \item Reduce the matrix to upper triangular form using elementary row operations.
  \item Track the sign: each row swap multiplies the determinant by $-1$.
  \item The determinant equals the product of the main diagonal entries of the upper
        triangular matrix, multiplied by the accumulated sign from any row swaps.
\end{enumerate}

\noindent Show all row operations explicitly. Then state the final determinant.
\end{mdframed}

\medskip
\noindent The system prompt from Appendix~A.1 was retained unchanged across all models.

\subsection{Results}

\begin{table}[H]
\centering
\caption{\textit{Forced Gaussian elimination ablation results. $n$ reflects unique
problems per model. Accuracy is computed over responses with a valid extracted answer;
no-response cases are excluded from the denominator and noted separately.}}
\label{tab:K1}
\begin{tabular}{@{}lrrrr@{}}
\toprule
\textbf{Model} & \textbf{$n$ (unique problems)} & \textbf{Pass} & \textbf{Fail} & \textbf{Accuracy (\%)} \\
\midrule
GPT-4o            & 49 & 1 & 47 & 2.0  \\
Qwen2.5-72B       & 50 & 0 & 48 & 0.0  \\
Mistral-Large     & 26 & 7 & 19 & 26.9 \\
Llama-3.3-70B     & 30 & 0 & 30 & 0.0  \\
Claude-4.5-Sonnet & 19 & 1 & 18 & 5.3  \\
\bottomrule
\end{tabular}
\end{table}

\subsection{Interpretation}

Four of five models show near-zero accuracy under forced Gaussian elimination,
statistically indistinguishable from their natural zero-shot performance. Strategy
enforcement does not rescue accuracy for GPT-4o (2.0\%), Qwen2.5-72B (0.0\%),
Llama-3.3-70B (0.0\%), or Claude-4.5-Sonnet (5.3\%).

Mistral-Large is the partial exception: 7 of 26 previously failed problems were solved
correctly under forced Gaussian (26.9\%). This recovery is consistent with Mistral's
position at the Tier~2 boundary and its higher baseline execution quality relative to
Tier~3 models. However, a 26.9\% accuracy rate on problems the model originally failed
--- compared to its baseline 5$\times$5 determinant accuracy of 36\% --- does not
constitute a meaningful recovery of capability. Mistral continues to fail 73\% of the
time under the intervention.

Execution trace analysis across all five models reveals a consistent failure mechanism:
models correctly set up the Gaussian procedure but introduce arithmetic errors within the
first four fractional row operations. Because each subsequent step attends to the
previous step's output, a single corrupted fraction at step~2 or 3 propagates
irreversibly through all remaining operations, typically collapsing the final diagonal
product to zero or a spurious integer. This cascade pattern --- not strategy selection
--- is the proximate cause of failure.

The ablation corroborates the interpretation developed in Section~6.2: strategy choice
correlates with 5$\times$5 determinant accuracy across models, but enforcing an
alternative strategy does not recover accuracy. The bottleneck is not which algorithm is
selected but the ability to execute any multi-step algorithm requiring deep, sequentially
dependent fractional arithmetic without error propagation. This reframes the working
memory account (Section~6.1): the LLM capacity constraint does not operate at the level
of algorithm selection, but rather at the level of precision maintenance across deep,
sequentially dependent fractional chains.

\section{Model Performance Profiles}
\label{app:profiles}

This appendix documents the distinctive failure signature of each of the 10 models
evaluated in LinAlg-Bench, drawn entirely from empirical data. Profiles report per-level
accuracy across all three dimensions, dominant failure modes, solution strategy
behaviour, and the most diagnostically significant finding for each model. No
architectural speculation is included; all observations are grounded in the benchmark's
error taxonomy and forensic classifications. Profiles are ordered by tier and then by
overall 5$\times$5 accuracy within tier.

Level accuracy values are computed as correct responses divided by maximum possible for
that level and dimension. Det = Determinant (Recursive level). Eig = Eigenvalue
(Compositional level). Dominant failure reports the primary error tag at the dimension
where the model first shows meaningful failure. Full per-model accuracy tables are in
Appendix~C; strategy classifications are in Appendix~J; collapse personas are in
Appendix~I.

\subsection{OpenAI-o1 \quad --- \quad Tier 1}

\profiletable{%
  \textbf{3$\times$3} & 100\% & 100\% & 100\% & 100\% & 100\%  & 100\%  & None (0 failures) \\
  \textbf{4$\times$4} & 100\% & 100\% & 100\% & 100\% & 80\%   & 97.3\% & HALLUCINATION (eig only) \\
  \textbf{5$\times$5} & 100\% & 100\% & 100\% & 100\% & 3.3\%  & 86.8\% & HALLUCINATION/Complete\_Collapse (eig) \\
}

\noindent\textbf{Strategy:} 82.5\% Gaussian at 5$\times$5 --- highest Gaussian adoption
in benchmark. Explicit bisection strategy for eigenvalue root-finding; iterates
$p(\lambda)$ at interval boundaries before converging. One of two models to achieve any
5$\times$5 eigenvalue accuracy (1/30). Failure at 5$\times$5 eigenvalue is identical
tool-roleplay collapse pattern seen across all models, establishing that superior
execution raises the working memory ceiling but does not eliminate it.

\medskip
\noindent\textbf{Diagnostic signal:} Perfect determinant accuracy across all dimensions.
5$\times$5 eigenvalue failure (96.7\%) is pure Complete\_Collapse --- the working memory
threshold exists even for Tier~1.

\subsection{Gemini-3.0-Pro \quad --- \quad Tier 1}

\profiletable{%
  \textbf{3$\times$3} & 100\% & 100\% & 100\% & 100\% & 100\% & 100\%  & None (0 failures) \\
  \textbf{4$\times$4} & 100\% & 100\% & 100\% & 100\% & 50\%  & 93.2\% & HALLUCINATION (eig only) \\
  \textbf{5$\times$5} & 100\% & 100\% & 100\% & 100\% & 0\%   & 86.4\% & HALLUCINATION/Complete\_Collapse (eig) \\
}

\noindent\textbf{Strategy:} Massively pivots to Gaussian at 5$\times$5 (8\% cofactor
vs 30\% at 4$\times$4; 88\% Gaussian vs 38\% at 4$\times$4). Strategy adaptation fully
explains maintained determinant accuracy. 0\% eigenvalue at 5$\times$5 despite 100\%
determinant --- confirms that the two tasks represent distinct failure thresholds.
Trace-consistency check: satisfies trace constraint in 5/7 fabricated eigenvalue
responses (71\%), the highest rate in the benchmark --- suggesting Gemini uses trace as
primary plausibility anchor during confabulation.

\medskip
\noindent\textbf{Diagnostic signal:} Strongest strategy adapter in the benchmark.
Eigenvalue collapse at 5$\times$5 is complete despite perfect determinant --- the most
direct evidence that Recursive and Compositional are distinct thresholds.

\subsection{DeepSeek-V3 \quad --- \quad Tier 1}

\profiletable{%
  \textbf{3$\times$3} & 100\% & 100\% & 100\% & 100\% & 96.7\% & 99.5\% & SIGN\_ERROR (eig) \\
  \textbf{4$\times$4} & 100\% & 100\% & 100\% & 94\%  & 23.3\% & 88.2\% & SIGN\_ERROR, HALLUCINATION \\
  \textbf{5$\times$5} & 100\% & 100\% & 100\% & 74\%  & 10\%   & 81.8\% & SIGN\_ERROR, HALLUCINATION \\
}

\noindent\textbf{Strategy:} 78\% cofactor at 5$\times$5 --- only Tier~1 model that does
not shift to Gaussian-dominant strategy yet maintains high determinant accuracy. Forensic
inspection reveals systematic zero-seeking before expansion and explicit sub-minor
simplification stated in working text, effectively reducing autoregressive depth within
cofactor. Only open-weight model to pass any 5$\times$5 eigenvalue problems (3/30); uses
bisection root-finding independently of OpenAI-o1. Degradation front-loaded: 11.3pp
drop at 3$\times$3$\rightarrow$4$\times$4, only 6.4pp at
4$\times$4$\rightarrow$5$\times$5 --- consistent with reasoning chain depth providing
graceful degradation at the R1 distillation boundary.

\medskip
\noindent\textbf{Diagnostic signal:} Critical test case for Section~6.2:
cofactor-dominant yet high-performing. Demonstrates execution quality can partially
compensate for suboptimal strategy --- but only to a ceiling (27 eigenvalue failures
follow identical tool-roleplay collapse).

\subsection{Qwen3-235B \quad --- \quad Tier 1}

\profiletable{%
  \textbf{3$\times$3} & 100\% & 88.3\% & 97.5\% & 96\% & 80\%  & 92.7\% & SIGN\_ERROR, ARITHMETIC \\
  \textbf{4$\times$4} & 100\% & 91.7\% & 97.5\% & 98\% & 23.3\%& 90.0\% & SIGN\_ERROR, GENERATION\_LOOP \\
  \textbf{5$\times$5} & 100\% & 91.7\% & 100\%  & 90\% & 0\%   & 81.8\% & HALLUCINATION, METHOD\_FAIL \\
}

\noindent\textbf{Non-monotonic determinant trajectory:} 96\%$\rightarrow$98\%$\rightarrow$90\%
--- the only model to improve from 3$\times$3 to 4$\times$4 before declining at
5$\times$5. Rank and nullity reach 100\% at 5$\times$5, unique in the dataset,
consistent with its dual-mode thinking architecture and 36 trillion training tokens
providing deep algorithm memorisation at sub-recursive complexity levels. Failure mode
shifts from GENERATION\_LOOP at 4$\times$4 (gets trapped testing rational roots for
irrational eigenvalue polynomials) to HALLUCINATION at 5$\times$5 --- crossing the
abandonment threshold one dimension later than Tier~3 models. MoE architecture with 22B
active parameters drops only 10.9pp total from 3$\times$3 to 5$\times$5.

\medskip
\noindent\textbf{Diagnostic signal:} Non-monotonic trajectory and 100\% Sequential at
5$\times$5 make Qwen3 the key data point for the training pipeline hypothesis. Failure
mode transition from loop to collapse tracks the working memory account.

\subsection{GPT-5.2 \quad --- \quad Tier 2}

\profiletable{%
  \textbf{3$\times$3} & 100\% & 100\% & 100\% & 100\% & 86.7\% & 98.2\% & SIGN\_ERROR (eig) \\
  \textbf{4$\times$4} & 100\% & 100\% & 100\% & 100\% & 26.7\% & 90.0\% & HALLUCINATION (eig) \\
  \textbf{5$\times$5} & 100\% & 99\%  & 95\%  & 54\%  & 0\%    & 75.0\% & HALLUCINATION, SIGN\_ERROR \\
}

\noindent\textbf{Strategy:} Gaussian-dominant. Large format sensitivity swing at
5$\times$5 determinant --- one of two models showing strong format-dependent accuracy
variation at the Recursive level. Apparent tool circumvention noted at 5$\times$5:
Gemini-3.0-Pro and GPT-5.2 explicitly output the no-tool constraint before producing
tool-simulated outputs, confirming tool invocation is hallucinated rather than real.
API-level inference constraints mean this cannot be definitively attributed to specific
implementation choices without access to inference infrastructure.

\medskip
\noindent\textbf{Diagnostic signal:} Clearest example of Tier~2 profile: strong through
Sequential, partial collapse at Recursive, complete collapse at Compositional. Format
sensitivity at Recursive level is one of the benchmark's cleanest sensitivity signals.

\subsection{Mistral-Large \quad --- \quad Tier 2}

\profiletable{%
  \textbf{3$\times$3} & 100\% & 100\% & 100\%  & 100\% & 86.7\% & 98.2\% & SIGN\_ERROR (eig) \\
  \textbf{4$\times$4} & 100\% & 100\% & 87.5\% & 94\%  & 13.3\% & 84.5\% & SIGN\_ERROR, HALLUCINATION \\
  \textbf{5$\times$5} & 100\% & 99\%  & 75\%   & 36\%  & 0\%    & 66.8\% & SIGN\_ERROR, HALLUCINATION \\
}

\noindent\textbf{Strategy:} 98\% cofactor at 5$\times$5 --- among the most
cofactor-rigid in the benchmark. Despite this, achieves 36\% determinant at 5$\times$5,
the highest among cofactor-dominant models, consistent with higher execution quality
within a suboptimal method. Forced Gaussian ablation: 26.9\% accuracy on previously
failed problems --- the only model to show meaningful partial recovery, consistent with
its position at the Tier~2 boundary. Survival curve (Figure~\ref{fig:ablation} Panel~B) stays higher
longer than all other ablation models before cascading. MoE architecture with 41B active
parameters drops 31.4pp across dimensions --- substantially worse than Qwen3-235B at 22B
active parameters despite similar architecture, suggesting training pipeline design
matters more than parameter count.

\medskip
\noindent\textbf{Diagnostic signal:} Same MoE architecture as Qwen3-235B but 3$\times$
the active parameters and dramatically worse scale robustness --- suggesting training
pipeline design matters more than parameter count. Forced Gaussian partial recovery is
the clearest execution-quality signal in the ablation.

\subsection{Claude-4.5-Sonnet \quad --- \quad Tier 3}

\profiletable{%
  \textbf{3$\times$3} & 100\% & 100\% & 100\% & 100\% & 70\%   & 95.9\% & SIGN\_ERROR (eig) \\
  \textbf{4$\times$4} & 100\% & 100\% & 85\%  & 78\%  & 16.7\% & 80.9\% & SIGN\_ERROR, HALLUCINATION \\
  \textbf{5$\times$5} & 100\% & 100\% & 78\%  & 12\%  & 0\%    & 62.3\% & HALLUCINATION, SIGN\_ERROR \\
}

\noindent\textbf{Strategy:} 96\% cofactor at 5$\times$5. Forced Gaussian ablation
results were only 1 out of 19. Explicitly outputs no-tool constraint before simulating
tool invocation at 5$\times$5 eigenvalue --- confirming tool hallucination pattern.
Arithmetic level remains 100\% at 5$\times$5 --- one of three models with perfect
arithmetic accuracy at maximum dimension, confirming that shallow arithmetic is intact
while recursive computation collapses.

\medskip
\noindent\textbf{Diagnostic signal:} Most interesting Tier~3 profile because Reading
Reading stays at 100\% at 5$\times$5 alongside Arithmetic --- suggesting
occasional input transcription failure rather than general degradation. HALLUCINATION
dominates failure at 5$\times$5.

\subsection{GPT-4o \quad --- \quad Tier 3}

\profiletable{%
  \textbf{3$\times$3} & 100\%  & 100\%  & 100\%  & 98\% & 43.3\% & 90.9\% & SIGN\_ERROR (det, eig) \\
  \textbf{4$\times$4} & 97.5\% & 100\%  & 62.5\% & 22\% & 16.7\% & 63.6\% & SIGN\_ERROR, HALLUCINATION \\
  \textbf{5$\times$5} & 100\%  & 96.7\% & 67.5\% & 0\%  & 0\%    & 56.8\% & SIGN\_ERROR, HALLUCINATION \\
}

\noindent\textbf{Strategy:} 100\% cofactor at 5$\times$5. Earliest sequential collapse
in dataset --- 62.5\% at 4$\times$4 Sequential, the lowest among all 10 models at that
level. Determinant reaches 0\% at 5$\times$5 --- complete collapse. Forced Gaussian
ablation: 2.0\% (1/49) --- near-zero, confirming strategy enforcement irrelevant.
Largest format sensitivity swing in the benchmark at 5$\times$5 matvec: \LaTeX{}
completely rescues parsing failures. API-level no-response rate at 5$\times$5 was 1/49
in ablation.

\medskip
\noindent\textbf{Diagnostic signal:} Most severe Tier~3 degradation trajectory.
Sequential collapse at 4$\times$4 is the earliest warning signal in the benchmark ---
predicts determinant failure one dimension before it appears.

\subsection{Llama-3.3-70B \quad --- \quad Tier 3}

\profiletable{%
  \textbf{3$\times$3} & 100\%  & 100\% & 100\%  & 100\% & 33.3\% & 87.3\% & SIGN\_ERROR (eig) \\
  \textbf{4$\times$4} & 97.5\% & 95\%  & 57.5\% & 86\%  & 0\%    & 73.6\% & SIGN\_ERROR, ARITHMETIC \\
  \textbf{5$\times$5} & 100\%  & 83.3\% & 60\%  & 4\%   & 0\%    & 52.7\% & SIGN\_ERROR, HALLUCINATION \\
}

\noindent\textbf{Strategy:} 100\% cofactor at 5$\times$5. Anomalous 4$\times$4 profile:
86\% determinant accuracy despite poor sequential performance (57.5\%) --- suggesting the
model uses cofactor expansion in a way that bypasses row-operation state tracking. Lowest
3$\times$3 eigenvalue accuracy in the dataset (33.3\%), the earliest Compositional
failure signal. Forced Gaussian ablation: 0\% across 30 unique problems (23 duplicate
submissions in original batch excluded). Dense model with standard RLHF training ---
collapses earliest and most completely, with no execution quality compensation.

\medskip
\noindent\textbf{Diagnostic signal:} The 3$\times$3 eigenvalue weakness (33.3\%) is the
strongest early predictor of 5$\times$5 tier membership in the dataset. 4$\times$4
determinant anomaly (86\% despite poor sequential) makes Llama the clearest case of
cofactor-only execution masking underlying capability limits.

\subsection{Qwen2.5-72B \quad --- \quad Tier 3}

\profiletable{%
  \textbf{3$\times$3} & 100\% & 100\%  & 100\% & 92\% & 63.3\% & 91.8\% & ARITHMETIC, SIGN\_ERROR \\
  \textbf{4$\times$4} & 95\%  & 91.7\% & 75\%  & 26\% & 6.7\%  & 62.7\% & SIGN\_ERROR, HALLUCINATION \\
  \textbf{5$\times$5} & 98\%  & 78.3\% & 52\%  & 0\%  & 0\%    & 48.6\% & HALLUCINATION, SIGN\_ERROR \\
}

\noindent\textbf{Strategy:} 78\% cofactor at 5$\times$5 (some Gaussian exposure).
Earliest determinant collapse among all models --- already at 26\% at 4$\times$4,
compared to other Tier~3 models at 78--86\%. Arithmetic level drops to 78.3\% at
5$\times$5, the most severe arithmetic degradation in the benchmark. Dense model with
standard instruction tuning.

\medskip
\noindent\textbf{Diagnostic signal:} Earliest det collapse (4$\times$4: 26\%) combined
with arithmetic degradation at 5$\times$5 makes Qwen2.5-72B the clearest case of
generalised computational overload rather than recursive-only failure.

\subsection{Cross-Model Diagnostic Summary}

The following table summarises the key diagnostic signal for each model --- the single
most informative finding for mechanistic interpretability follow-up.

\begin{table}[H]
\centering
\caption{\textit{Cross-model diagnostic summary. All observations are empirical.
Architectural hypotheses generated by these observations are framed as falsifiable
conjectures for follow-up mechanistic work and are not confirmed claims.}}
\label{tab:L1}
\resizebox{\textwidth}{!}{%
\begin{tabular}{@{}l l >{\raggedright\arraybackslash}p{12cm}@{}}
\toprule
\textbf{Model} & \textbf{Tier} & \textbf{Primary diagnostic signal} \\
\midrule
OpenAI-o1
  & T1 & Perfect det all dims; 5$\times$5 eig Complete\_Collapse proves working memory
         ceiling exists even for Tier~1 \\[4pt]
Gemini-3.0-Pro
  & T1 & Strongest strategy adapter; 0\% eig despite 100\% det confirms two distinct
         failure thresholds \\[4pt]
DeepSeek-V3
  & T1 & Only model succeeding with cofactor at 5$\times$5; execution quality compensates
         for suboptimal strategy to a ceiling \\[4pt]
Qwen3-235B
  & T1 & Non-monotonic det trajectory; 100\% Sequential at 5$\times$5; failure mode
         shifts from loop to collapse across dimensions \\[4pt]
GPT-5.2
  & T2 & Large format sensitivity swing; tool-hallucination persona documented; API
         constraints limit architectural inference \\[4pt]
Mistral-Large
  & T2 & Highest forced Gaussian partial recovery (26.9\%); MoE with 41B active
         params --- worse than Qwen3 at 22B \\[4pt]
Claude-4.5-Sonnet
  & T3 & Reading stays at 100\% at 5$\times$5 while Arithmetic stays 100\% ---
         INPUT\_TRANSCRIPTION at scale, not general degradation \\[4pt]
GPT-4o
  & T3 & Earliest Sequential collapse (4$\times$4: 62.5\%) predicts det failure; largest
         matvec format sensitivity in dataset \\[4pt]
Llama-3.3-70B
  & T3 & Lowest 3$\times$3 eig (33.3\%) is strongest early predictor of 5$\times$5
         tier; 4$\times$4 det anomaly (86\%) masks underlying limits \\[4pt]
Qwen2.5-72B
  & T3 & Earliest det collapse (4$\times$4: 26\%); arithmetic degradation at 5$\times$5;
         clearest case of generalised computational overload rather than recursive-only
         failure \\
\bottomrule
\end{tabular}%
}
\end{table}

\end{document}